\documentclass[10pt,twocolumn,letterpaper]{article}
\pdfoutput=1
\usepackage{iccv}
\usepackage{times}
\usepackage{epsfig}
\usepackage{graphicx}
\usepackage{amsmath}
\usepackage{amssymb}
\usepackage{ulem}
\usepackage{multirow}
\usepackage{subcaption}
\usepackage{algorithm}
\usepackage{algorithmic}

\usepackage[breaklinks=true,bookmarks=false]{hyperref}

\iccvfinalcopy % *** Uncomment this line for the final submission

\def\iccvPaperID{****} % *** Enter the ICCV Paper ID here

\begin{document}

%%%%%%%%% TITLE
\title{ Dynamic Knowledge Distillation With Noise Elimination for RGB-D Salient Object Detection}

\author{Guangyu Ren\\
Imperial College London\\
{\tt\small g.ren19@imperial.ac.uk}

\and
Yinxiao Yu\\
{\tt\small yz11010604@gmail.com}

\and
Hengyan Liu\\
Imperial College London\\
{\tt\small hengyan.liu15@imperial.ac.uk}

\and
Tania Stathaki\\
Imperial College London\\
{\tt\small t.stathaki@imperial.ac.uk}
}

\maketitle
%\thispagestyle{empty}

%%%%%%%%% ABSTRACT
\begin{abstract}
RGB-D salient object detection (SOD) demonstrates its superiority on detecting in complex environments due to the additional depth information introduced in the data. Inevitably, an independent stream is introduced to extract features from depth images, leading to extra computation and parameters. This methodology sacrifices the model size to improve the detection accuracy which may impede the practical application of SOD problems. To tackle this dilemma, we propose a dynamic distillation method along with a lightweight structure, which significantly reduces the computational burden while maintaining validity. This method considers the factors of both teacher and student performance within the training stage and dynamically assigns the distillation weight instead of applying a fixed weight on the student model. We also investigate the issue of RGB-D early fusion strategy in distillation and propose a simple noise elimination method to mitigate the impact of distorted training data caused by low quality depth maps. Extensive experiments are conducted on five public datasets to demonstrate that our method can achieve competitive performance with a fast inference speed (136FPS) compared to 10 prior methods.
\end{abstract}

%%%%%%%%% BODY TEXT
\section{Introduction}

\begin{figure}[h]
%\minipage{0.5\textwidth}
  \centering
  \includegraphics[width=\linewidth]{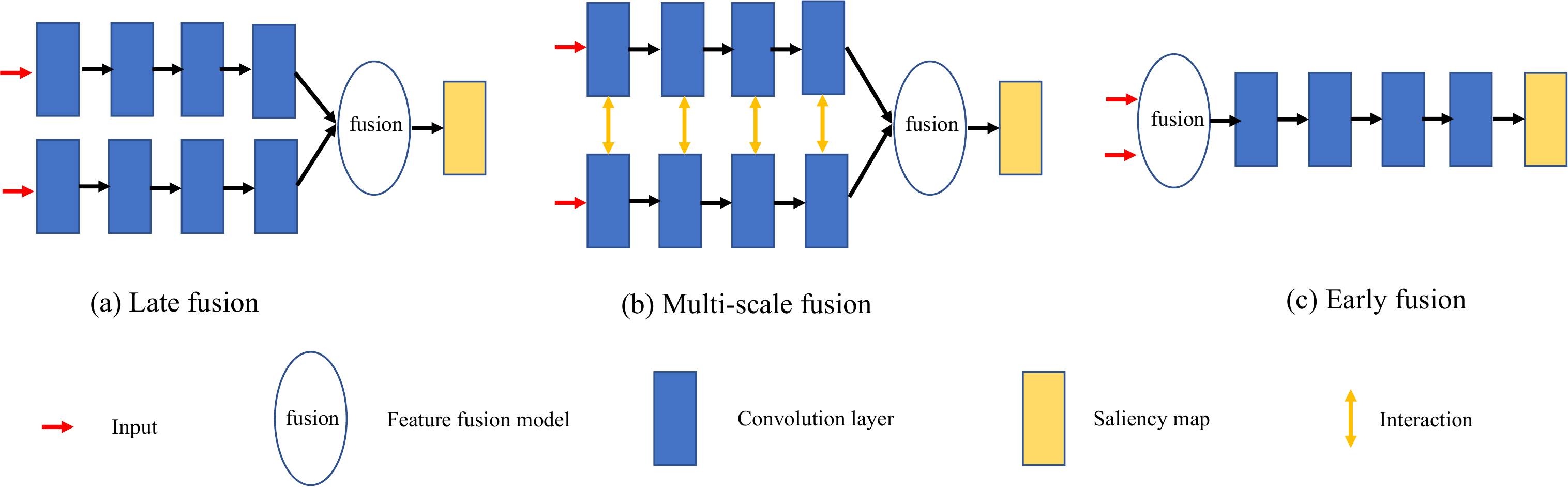}
  %(a) 
%\endminipage\hfill

\caption{Comparison of three fusion strategies.}
\label{fig:fusion ways}
\end{figure}

Salient object detection (SOD) aims at locating prominent objects in a given scenario under consideration. In recent years SOD has attracted significant attention and substantial progress has been demonstrated in the field. The object detection task can be treated as a pre-processing methodology that can be subsequently used in diverse fields, such as image understanding~\cite{zhu2014unsupervised}, video detection and
segmentation~\cite{fan2019shifting}, semantic segmentation~\cite{shimoda2016distinct}, object tracking~\cite{mahadevan2009saliency}, person re-identification~\cite{zhao2016person} and others. However, due to the complicated real-life scenarios, RGB-based SOD still fails in generating satisfactory prediction maps. In order to overcome this issue and obtain better detection performance in complex scenarios, depth images along with an independent network have been introduced to provide supplementary information. Specifically, Figure \ref{fig:fusion ways} illustrates three fusion methods in RGB-D based SOD. Existing state-of-the-art methods mainly adopt late fusion or multi-scale fusion and focus on designing feature-enhanced modules and complicated feature-fusion modules,  which indeed improve the overall detection results. However, due to the processing of high volume of information the models tend to become extremely complicated, an issue which weakens the practicality of SOD using RGB-D data. The early fusion in Figure~\ref{fig:fusion ways}(c)  integrates the separate inputs into a unified representation before the feature extraction process. It provides an alternative strategy to lightens the model but suffers from the noise issue caused by low-quality depth information. That motivates us to seek to explore the potential of early fusion and compress the model size for SOD and while maintaining high detection accuracy. 
%Our final model achieves a good balance between accuracy and model size on widely used benchmarks as shown in Figure \ref{fig:balance between accuracy and model size} 

Recently, Knowledge Distillation (KD) has been proposed~\cite{hinton2015distilling} to transfer knowledge from a large model to a smaller one. The main idea is that a small student model mimics a cumbersome model, namely a so-called teacher model, to achieve a competitive performance.  The cumbersome network has larger knowledge capacity than smaller models, but this capacity may not be utilized for its full potential. In other words, a lightweight network can reach a similar performance to a cumbersome network by KD without increasing the number of parameters. Similar to human behaviours, this teacher-student learning process can be implemented by a simple and effective way, which forces the student model to directly learn the final prediction of the teacher model.

\begin{figure}[h!]
\minipage{0.23\textwidth}
  \centering
  \includegraphics[width=\linewidth]{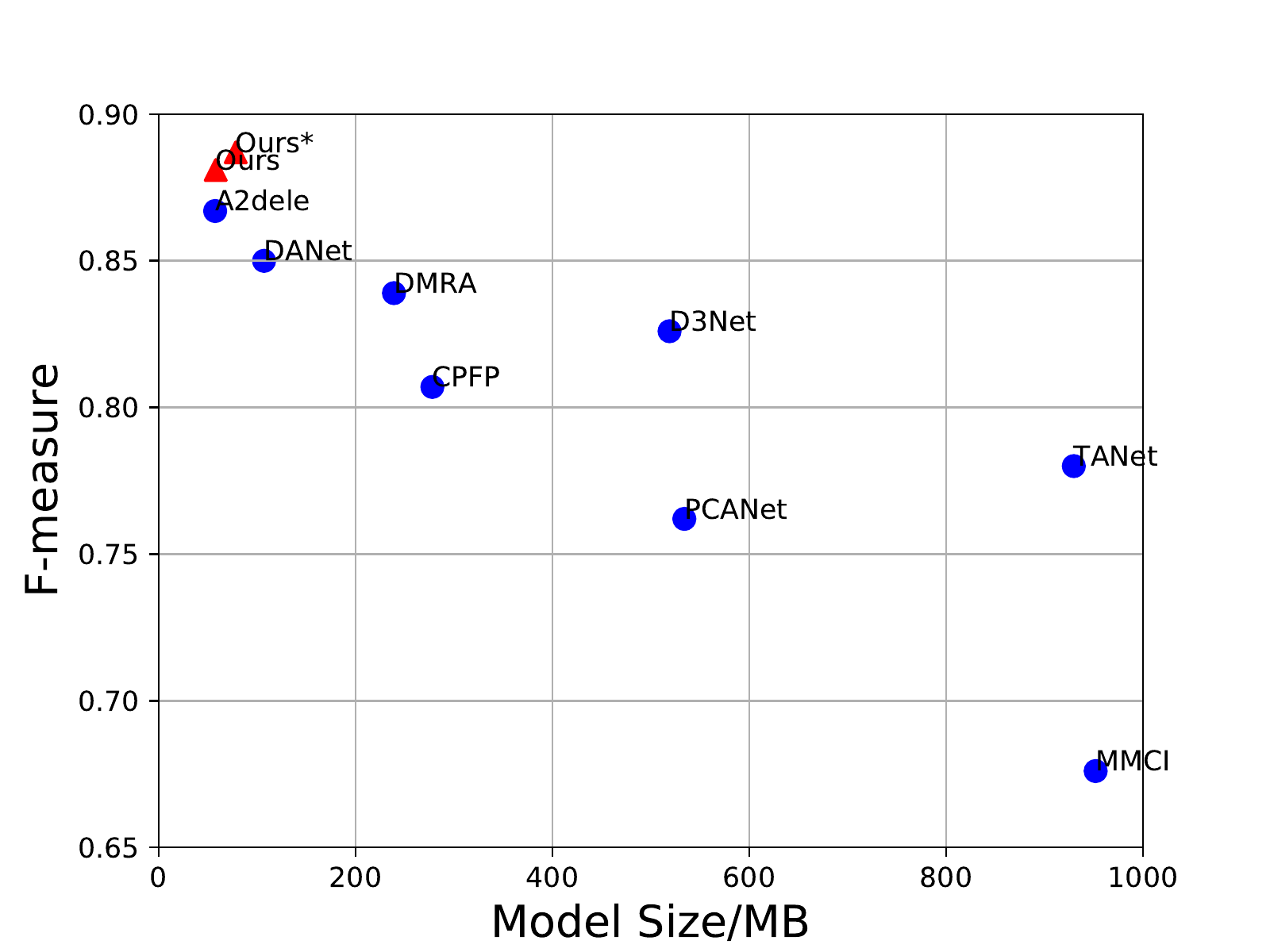}
  (a) NLPR
\endminipage\hfill
\minipage{0.23\textwidth}%
  \centering
  \includegraphics[width=\linewidth]{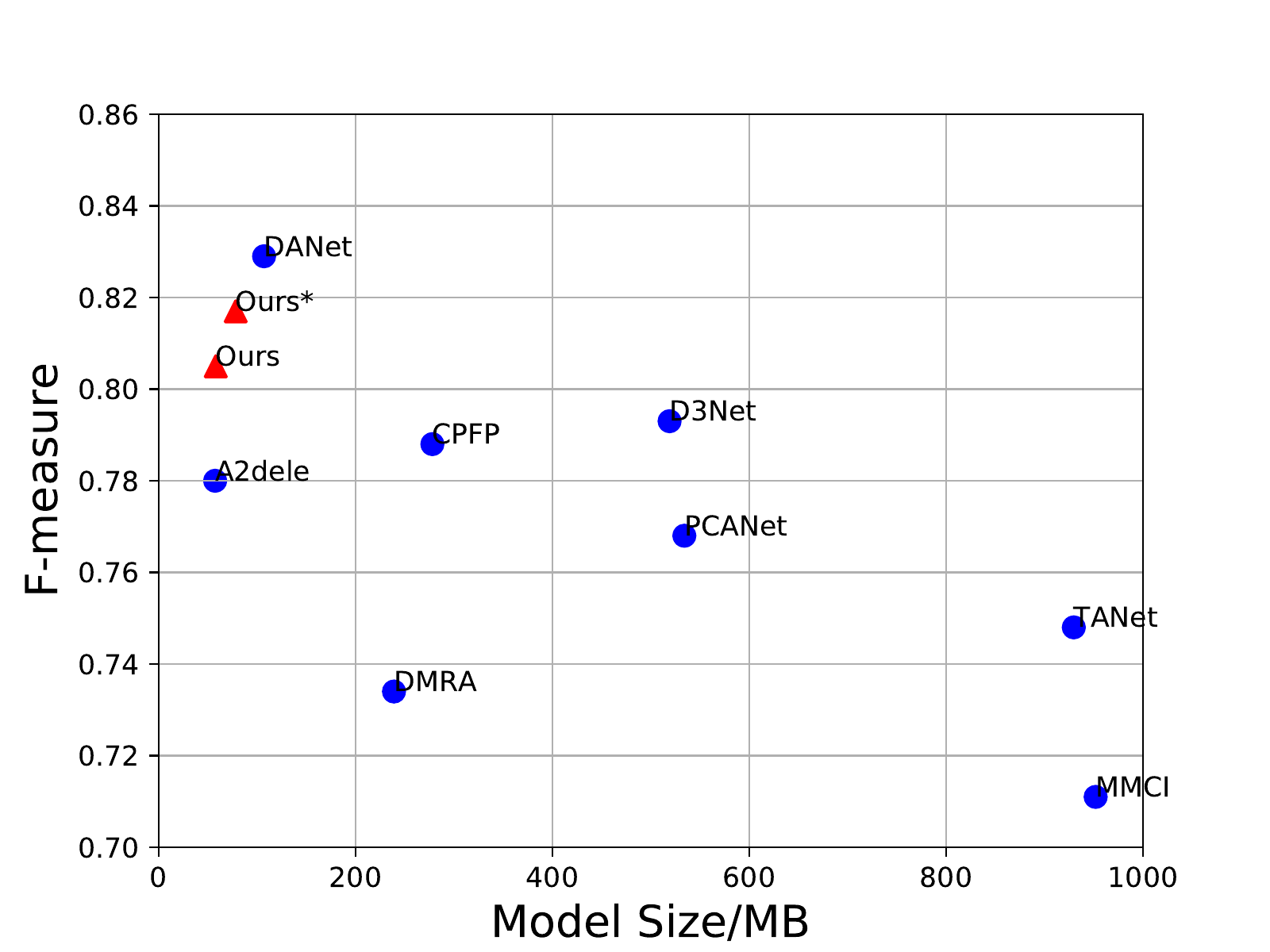}
  (b) SIP
\endminipage\hfill

\caption{Weighted F-measure and model size on NLPR and SIP datasets. Our lightweight model with the simple early fusion can achieve satisfactory detection results over different datasets. $Ours$ and $Ours*$ indicate simple VGG16-based and VGG19-based FPN respectively.}

\label{fig:balance between accuracy and model size}
\end{figure}

KD has been applied in a range of machine learning applications. Zhang~\cite{zhang2019training} utilizes the knowledge distillation to RGB saliency detection and proposes an efficient model by reducing the number of channels. Piao~\cite{piao2020a2dele} explores the cross-modal distillation on RGB-D data and uses an adaptive weight to distil the depth knowledge from the teacher model. Nevertheless, both adjust their student networks according to the teacher networks and distillation strategies. Besides, the adaptive distillation~\cite{piao2020a2dele} is proposed for the cross-modal distillation and only considers the performance of the teacher model, which limits the utilization of this KD method.

In order to tackle the above issues from a new perspective, we use a concise framework based on the early fusion strategy for RGB-D based SOD and propose a dynamic knowledge distillation weight to help the model pay more attention on hard samples by considering both teacher and student performance. We also investigate the issue of RGB-D early fusion strategy in distillation and propose a simple noise elimination method to mitigate the impact of distorted training data caused by low quality depth maps. Combing these two methods can lead to a reasonable distillation strategy for RGB-D saliency detection. Our final model achieves a good balance between accuracy and model size on widely used benchmarks as shown in Figure \ref{fig:balance between accuracy and model size}. In a nutshell, our main contributions can be summarised as follows:
\begin{itemize}

\item We propose a novel dynamic distillation strategy, which can adaptively assign the distillation weight which simultaneously considers detection performance of the teacher and student networks within the training stage. As a result, the final model can pay more attention on hard samples and improve the overall performance.

\item We propose a noise elimination method by taking full merit of knowledge prior from the teacher network to alleviate the impact of depth maps with
 low quality. The student network can take benefit from this method without increasing extra parameters and computations.

\item We adopt a single stream for RGB-D SOD in order to bypass the depth network and avoid designing a complicated model. This single stream achieves competitive performance by only using VGG16 (57.9MB) and VGG19 (78.2MB), which are more applicable for practical use. Extensive experimental results on five benchmarks demonstrate that our methods can achieve competing performance within a fast lightweight architecture.

% \item We show that a teacher with higher performance usually benefits from knowledge distillation on salient object detection. This phenomenon is same as object detection and different from image classification.
\end{itemize}

\begin{figure*}[h]
%\minipage{0.5\textwidth}
  \centering
  \includegraphics[width=0.8\linewidth]{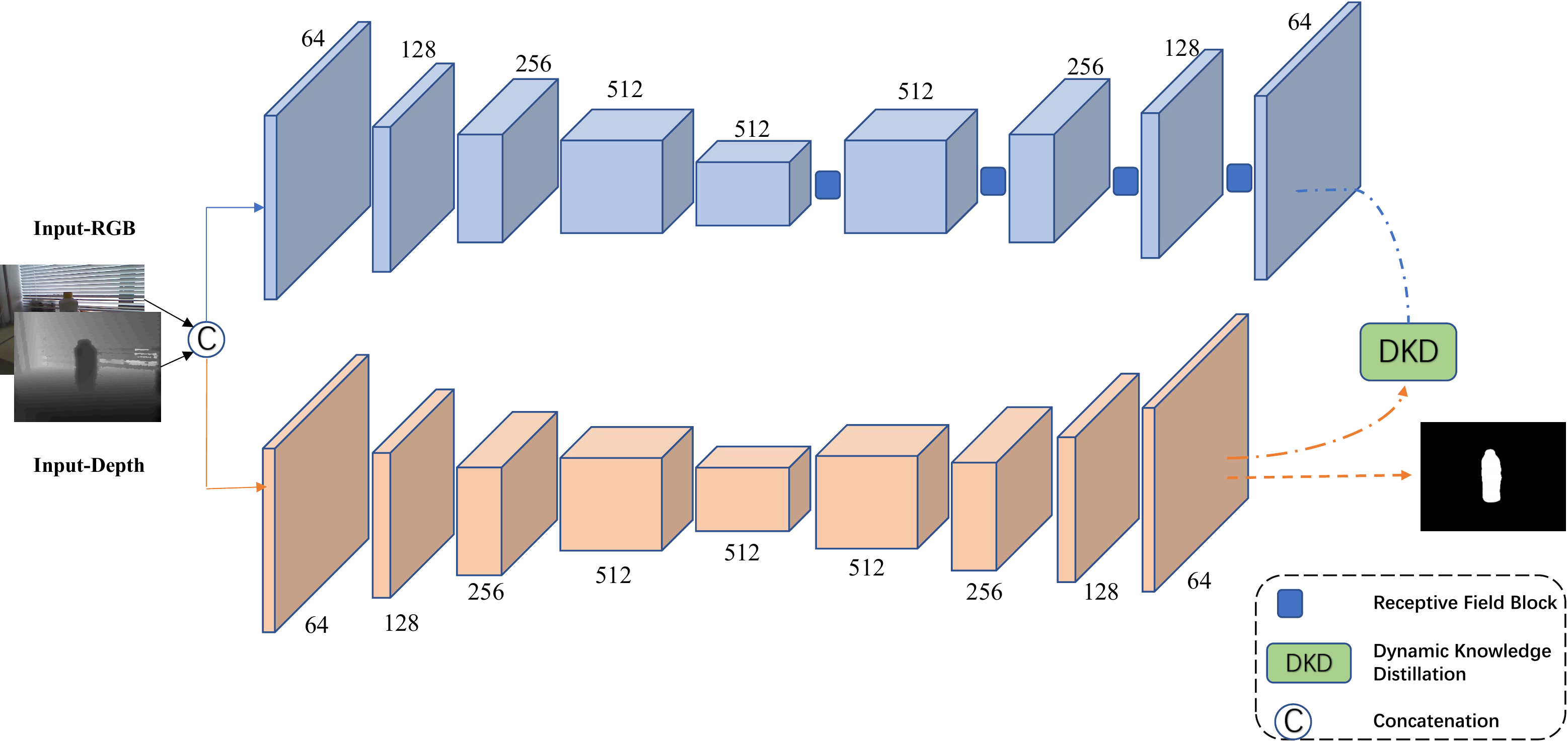}
  %(a) 
%\endminipage\hfill

\caption{Our framework consists of two stages. We adopt a cumbersome network as the teacher model and a feature pyramid network as the student model.}
\label{fig:overall_structure}
\end{figure*}

\section{Related Work}
\noindent\textbf{RGB-D Salient Object Detection.} RGB-D based SOD has obtained increasing attention in order to handle object detection tasks in complicated environments. Depth information is firstly introduced by~\cite{lang2012depth}, where they model the distribution of depth-induced saliency by using Gaussian mixture models. Zhao~\cite{zhao2019contrast} proposes a feature-enhanced module and a contrast-enhanced net, which augments the contrast between the foreground and background by fluid pyramid integration. Pang~\cite{pang2020hierarchical} adopts multi-scale fusion and proposes a dynamic dilated pyramid module with adaptive receptive fields, which is generated by densely integrating cross-modal features. Chen~\cite{chen2020progressively} constructs a lightweight depth stream and designs a refinement network, which is progressively stacked by guided residual blocks. This method can alternately alleviate the mutual degradation and refine predictions in a progressive way. These previous works focus on alleviating the impact of depth maps and enhancing the feature integration through delicate modules and networks.

\noindent\textbf{Knowledge Distillation.} Knowledge distillation was formally publicised by ~\cite{hinton2015distilling} in a teacher-student learning framework. This method proposes an effective way to compress model size and attempts to imitate the human beings’ learning mechanism. Cheng~\cite{cheng2020explaining} designs mathematical metrics to quantify and compare the methods of learning from teacher model and learning from raw data. They explain the superiority of knowledge distillation in three aspects. First, more reliable visual concepts can be learned through knowledge distillation. Second, knowledge distillation makes the model able to learn various concepts simultaneously. Furthermore, learning from KD can generate more stable optimization directions in the training phase. Recently, knowledge distillation has been used in SOD tasks. Zhang~\cite{zhang2019training} designs the student model by reducing the amount of channels and applies multi-scale knowledge distillation on the corresponding scales between teacher and student models. Piao~\cite{piao2020a2dele} applies cross-modal distillation on RGB-D based SOD and proposes an adaptive distiller to distil the depth information, which alleviates the impact of low-quality depth maps.

% Different from the above methods, our method bypasses the delicate network structures and complicated distillation processes. It considers the knowledge confidence in teacher and hard samples in student. The proposed method can be treated as a general distillation method and simply applied not only for SOD but for diverse tasks.

\section{Methodologies}

\subsection{Overview}
Existing methodologies for RGB-D SOD tend to build two-stream networks in order to process RGB and depth features separately. This two-stream design could improve detection performance but meanwhile introduces a large amount of parameters, which increases the complexity and reduces the practicality of models. Feature pyramid network (FPN)~\cite{lin2017feature} is an effective structure which utilizes multi-scale features in different resolutions to achieve detection tasks. In this work, we do not focus on designing networks and only adopt the classic FPN based on a VGG16 and a VGG19~\cite{simonyan2014vgg} as the student model. In order to obtain a stronger teacher model, we employ four receptive field blocks~\cite{liu2018rfb} in multi-scale layers to boost the detection performance. Considering different cross-modal fusion strategies, we choose the simple early fusion way which directly concatenates RGB images and depth images to form four-channel inputs. Similar to normal knowledge distillation, we transfer the probability distribution of the final layer from the teacher model to the student model by utilizing the so-called dynamic KD.

\subsection{Dynamic Knowledge Distillation
}

As mentioned above, Knowledge Distillation (KD) benefits the student model but the weight of knowledge transfer is still hand-designed. Piao~\cite{piao2020a2dele} proposes an adaptive weight for cross-modal distillation. However, in~\cite{piao2020a2dele} they only distill the depth information by considering the performance of teacher model. In our method, we consider both performances of teacher and student networks and combine these two factors as a dynamic weight for KD. 
% Besides, we show that the previous adaptive method is a special case of our method. 

Concretely, the accuracy of teacher model represents the detection performance which also indicates the confidence of knowledge. Inspired by IOU~\cite{mattyus2017iou} used in SOD, we design a dynamic factor $\alpha_{t}$ to modulate the correct knowledge which can be transferred from the teacher model as follows:

\begin{equation}
    \centering
     %\alpha_{t} = exp(-L_{CE}({P_{t},G}))
     \alpha_{t} = \dfrac {P_{t}\cdot G}{P_{t}+G - P_{t}\cdot G}
\end{equation}
where $P_{t}$ and $G$ represent the prediction of teacher model and the ground truth, respectively. $\alpha_{t}$ indicates the confidence of knowledge which can be transferred to the student model. Then, we propose another dynamic factor $\beta_{s}$ to show the degree of desired knowledge for the student model as follows:

\begin{equation}
    \centering
     \beta_{s} = 1-\dfrac {P_{s}\cdot G}{P_{s}+G - P_{s}\cdot G}
     %\beta_{s} = Sigmoid(1-\dfrac {P_{s}\cdot G}{P_{s}+G - P_{s}\cdot G})
\end{equation}
where $P_{s}$ represents the prediction of student model. This dynamic factor $\beta_{s}$ is error rate of the current training sample. In other words, knowledge distillation should also consider the current performance of student model. $\beta_{s}$ is inversely related to the accuracy between the output of student model and the ground truth. This indicates that hard samples which have large error rates need to learn more from the teacher model.
Therefore, we propose a simple and effective formulation to find a plausible distillation weight $\theta_{t,s}$:

\begin{equation}
    \centering
     \theta_{t,s} = tanh({\alpha_{t}}^{p} \cdot  {\beta_{s}}^{1-p})
\end{equation}
here $tanh$ is treated as a scale function:

\begin{equation}
    \centering
     tanh(x) = \dfrac {exp(x)- exp(-x)}{exp(x)+exp(-x)}
\end{equation}

More specifically, we define the $\theta_{t,s}$ by the weighted geometric mean of the knowledge confidence $\alpha_{t}$ from teacher and the knowledge demand $\beta_{s}$ from student. We define the hyper-parameter $p \in [0,1]$ to balance the ratio between the teacher and student networks. It is worth noting that large variation of $\theta_{t,s}$ leads to convergence issue in training phase. In this case, we further use a $tanh$ function to scale the $\theta_{t,s}$.  The overall loss function can be formulated as:

\begin{equation}
    \centering
     L_{dynamic} = \theta_{t,s} L_{KL}(P_{s},P_{t}) + (1-\theta_{t,s}) L_{CE}(P_{s},G)
\end{equation}
where $L_{KL}$ is the Kullback-Leibler divergence loss and $L_{CE}$ represents the cross-entropy loss. In the final network, we set the distillation temperature to 5 in $L_{KL}$ and $p=0.7$.

\subsection{Noises Elimination with the Dynamic Knowledge Distillation}

\begin{figure}
%\minipage{0.8\textwidth}
  \centering
  \includegraphics[width=0.9\linewidth]{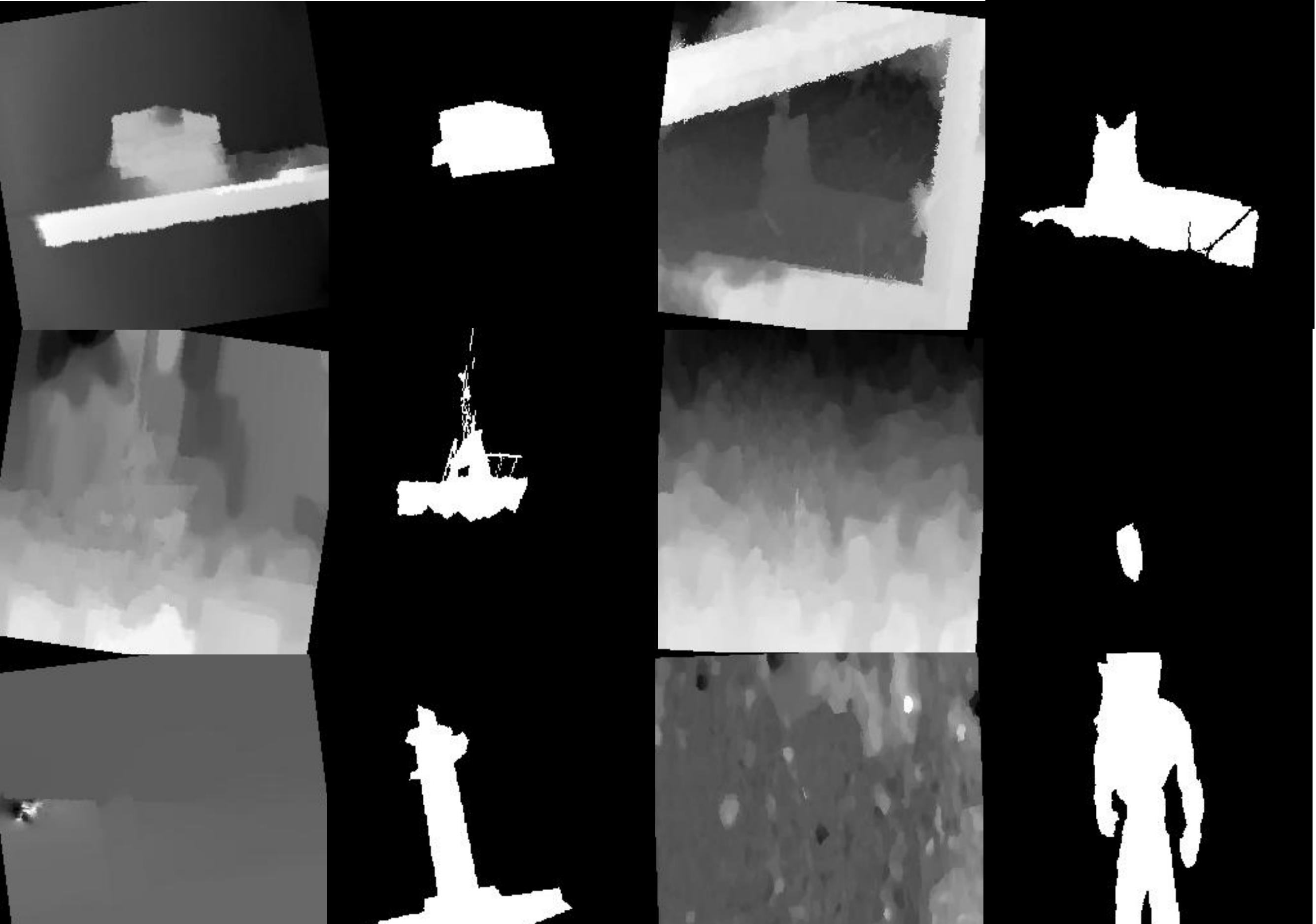}

\caption{ Low quality depths in training.}
\label{fig:low quality depth}
\end{figure}

As mentioned above, we simplify the procedure of knowledge distillation and the student network architecture in RGB-D task. Concretely, we only distill the final output distribution and abandon the depth stream by concatenating RGB and depth maps to form a four-channel input.
However, this fusion strategy suffers from the noise issue caused by low-quality depth information. As illustrated in Figure~\ref{fig:low quality depth}, we investigate the reasons that cause the distortion of depth maps: 1) besides the salient object, other objects in depth image dominate salient features; 2) low contrast between salient object and background in depth; 3) depth distortion caused by camera. Intuitively, training loss is supposed to reduce drastically if the training data is distorted. Therefore we propose that these depth maps can be treated as noises when combining with RGB maps and further set an accuracy threshold during knowledge distillation to control the impact of noises:

\begin{equation}
\theta_{t,s}=\left\{
\begin{array}{cl}
tanh({\alpha_{t}}^{p} \cdot  {\beta_{s}}^{1-p}) &  \alpha_{t} > threshold \\
\epsilon  &  Others \\
\end{array} \right.
\end{equation}where $\epsilon$ indicates a small weight which is set to 0.01 in this paper. $\alpha_{t}$ provides a knowledge prior from teacher network and indicates whether the depth distortion happens. Here $threshold$ is set to 0.5. Under this circumstance, the student model is able to know the useless training data when receiving knowledge from the teacher model.

Compared to considering one aspect or enforcing a fixed weight to the student model, our dynamic weight considers both the correctness of teacher's knowledge and the error of student network, which allows the student network to receive the knowledge according to the degree of difficulty of samples. $\theta_{t,s}$ varies little in the start of training phase. As for late stage of training, the student network is able to detect most simple scenarios except for some hard samples. Therefore, $\theta_{t,s}$ automatically assign to relatively bigger weights for hard samples which can be detect accurately in teacher network but student network. The noise elimination method takes full merit of the knowledge prior from teacher network and effectively reduce the negative impact of depth maps in low quality. Extensive experiments demonstrate in section 4 that this dynamic knowledge distillation could boost the detection performance without increasing extra parameters and model size. The process of the proposed methods is illustrated as algorithm \ref{algo:alg1}

.

\begin{algorithm}
	\caption{\label{algo:alg1} Dynamic Knowledge Distillation.
% 	reward for policy gradient.
	}
	\begin{algorithmic}[1]
		\REQUIRE
		$P_{t}$ is the prediction of teacher network,\\
		$P_{s}$ is the prediction of student network,\\
		$G$ is the corresponding ground truth.
		%\ENSURE policy parameters $\theta$ and intrinsic module parameters $\eta$.
		%\STATE \textbf{Init}: initialize the policy network $\pi_{\theta}$ and intrinsic network $\phi_{\eta}$.
% 		\FOR{Iteration \textbf{in} 1, 2, 3, ..., N}
		%\FOR{i = 5, 4, 3, 2, 1}
		\STATE {\textcolor{black} {\textbf{Stage 1:}}}Training the teacher network;
		\STATE $loss = L_{CE}(P_{t},G)$
		\STATE {\textcolor{black} {\textbf{Stage 2:}}}Training the student network;
		\STATE $\alpha_{t} = IOU(P_{t},G)$;
		\STATE $\beta_{s} = 1-IOU(P_{s},G)$;
		
		\IF{$\alpha_{t}> Threshold$}
            \STATE $\theta_{t,s} = tanh({\alpha_{t}}^{p} \cdot  {\beta_{s}}^{1-p})$
        
        \ELSE
            \STATE $\theta_{t,s} = 0.01$
            
        \ENDIF

		\STATE $loss = \theta_{t,s} L_{KL}(P_{s},P_{t}) + (1-\theta_{t,s}) L_{CE}(P_{s},G)$;
		
		%\ENDFOR
        %\ENDWHILE
        %\RETURN ${Output}^{1}$;
	\end{algorithmic}
	\label{algo:alg1}
\end{algorithm}

\begin{table*}[h]
\centering
\caption{Quantitative comparisons through the maximum of $F$-score $F_{\beta}$, S-score $S_{\alpha}$, E-score $E_{\theta}$, and error-score $M$, over five widely evaluated datasets. $Ours$ and $Ours*$ indicate simple VGG16-based and VGG19-based FPN repsectively. The best scores are highlighted in {\color{black} \textbf{bold}}.}
\label{table:overall quantitive in distillation}
\resizebox{\linewidth}{!}{
\begin{tabular}{cc|cccccccccc|c|c}
\hline
\hline
\multirow{2}{*}{} & {Metric}  & DF  & CTMF   & AFNet    & MMCI  & TANet  & DMRA   & CPFP   & D3Net  & A2dele & DANet &Ours   & Ours*    \\ \hline
\multirow{4}{*}{\rotatebox{90}{NJUD}~\rotatebox{90}{}}    & $F_{\beta}\uparrow$  & 0.789 & 0.857 & 0.804 & 0.868 & 0.888 & 0.896  & 0.890 &0.903 &0.905 &0.890 &0.928  & {\textcolor{black} {\textbf{0.934}}} \\
                         & $S_{\alpha}\uparrow$  & 0.735 & 0.849 & 0.772 & 0.859 & 0.878 & 0.885 & 0.878 & 0.895 &0.867&0.897  &0.916& {\textcolor{black} {\textbf{0.920}}} \\
                         & $E_{\theta}\uparrow$  & 0.818 & 0.866 & 0.847 & 0.882 & 0.909 & 0.920 & 0.900 & 0.901 &0.914&0.926 &0.949 & {\textcolor{black} {\textbf{0.952}}} \\
                         & $M\,\downarrow$  & 0.151 & 0.085 & 0.100 & 0.079 & 0.061 & 0.051 & 0.053 & 0.051 & 0.052 &0.046 & 0.032& {\textcolor{black} {\textbf{0.030}}} \\ \hline
\multirow{4}{*}{\rotatebox{90}{NLPR}~\rotatebox{90}{}}    & $F_{\beta}\uparrow$  & 0.752 & 0.841 & 0.816 & 0.841 & 0.876 & 0.888 & 0.884 & 0.904 &0.891 & 0.908  &0.922 & {\textcolor{black} {\textbf{0.930}}} \\
                         & $S_{\alpha}\uparrow$  & 0.769 & 0.860 & 0.799 & 0.856 & 0.886 & 0.898 & 0.884 & 0.906 & 0.889 & 0.908  &0.921 & {\textcolor{black} {\textbf{0.924}}} \\
                         & $E_{\theta}\uparrow$  & 0.840 & 0.869 & 0.884 & 0.872 & 0.926 & 0.942 & 0.920 & 0.934 & 0.937 & 0.945 &0.958  & {\textcolor{black} {\textbf{0.960}}} \\
                         & $M\,\downarrow$  & 0.110 & 0.056 & 0.058 & 0.059 & 0.041 & 0.031 & 0.038 & 0.034 & 0.031 & 0.031  &0.022 & {\textcolor{black} {\textbf{0.021}}} \\ \hline
\multirow{4}{*}{\rotatebox{90}{DES}~\rotatebox{90}{}}    & $F_{\beta}\uparrow$  & 0.625 & 0.865 & 0.775 & 0.839 & 0.853 & 0.906 & 0.882 & 0.917 & 0.897 & 0.916 &0.926  & {\textcolor{black} {\textbf{0.928}}} \\
                         & $S_{\alpha}\uparrow$  &0.685 & 0.863 & 0.770 &0.848 & 0.858 & 0.899 & 0.872 &0.904 & 0.883 & 0.905 &0.918  & {\textcolor{black} {\textbf{0.918}}} \\
                         & $E_{\theta}\uparrow$  & 0.806 & 0.911 & 0.874 & 0.904 & 0.919 & 0.944 & 0.927 & 0.956 & 0.918 & 0.961 &0.965  & {\textcolor{black} {\textbf{0.966}}} \\
                         & $M\,\downarrow$  & 0.131 & 0.055 & 0.068 & 0.065 & 0.046 & 0.030 & 0.038 & 0.030 & 0.030 & 0.028 &{\textcolor{black} {\textbf{0.022}}} & 0.023 \\ \hline
\multirow{4}{*}{\rotatebox{90}{LFSD}~\rotatebox{90}{}}    & $F_{\beta}\uparrow$  & 0.854 & 0.815 & 0.780 & 0.813 & 0.827 & {\textcolor{black} {\textbf{0.872}}} & 0.850 & 0.849 & 0.858 & - &0.865  & 0.862 \\
                         & $S_{\alpha}\uparrow$  & 0.786 & 0.796 & 0.738 &0.787 & 0.801 & 0.847 & 0.828 & 0.832 & 0.833 & - &0.834  & {\textcolor{black} {\textbf{0.839}}} \\
                         & $E_{\theta}\uparrow$  & 0.841 & 0.851 & 0.810 & 0.840 & 0.851 & {\textcolor{black} {\textbf{0.899}}} & 0.867 & 0.860 & 0.875 & - &0.883   & 0.883\\
                         & $M\,\downarrow$  & 0.142 & 0.120 & 0.133 & 0.132 & 0.111 & 0.076 & 0.088 & 0.099 & 0.077 & - &0.080 & 0.078 \\ \hline
\multirow{4}{*}{\rotatebox{90}{SIP}~\rotatebox{90}{}}    & $F_{\beta}\uparrow$  & 0.704 & 0.720 & 0.756 & 0.840 & 0.851 & 0.847 & 0.870 & 0.882 & 0.855 & {\textcolor{black} {\textbf{0.901}}} &0.872  & 0.882 \\
                     & $S_{\alpha}\uparrow$  & 0.653 & 0.716 & 0.720 & 0.833 & 0.835 & 0.800 & 0.850 &  0.864 & 0.828 & {\textcolor{black} {\textbf{0.878}}} &0.855  & 0.865 \\
                         & $E_{\theta}\uparrow$  & 0.794 & 0.824 &0.815 & 0.886 & 0.894 & 0.858 &  0.899 & 0.903 & 0.890 &{\textcolor{black} {\textbf{0.914}}} &0.908 & {\textcolor{black} {\textbf{0.914}}} \\
                         & $M\,\downarrow$  & 0.185 & 0.139 & 0.118 & 0.086 & 0.075 & 0.088 & 0.064 & 0.063 & 0.070  & {\textcolor{black} {\textbf{0.054}}} &0.060 & 0.056 \\ \hline

\hline
\end{tabular}%}
}
\end{table*}

\begin{table*}
\centering
\resizebox{\linewidth}{!}{
\begin{tabular}{c|c|c|c|c|c|c|c|c|c}
\hline
Method   & MMCI & TANet & PCANet & D3Net& CPFP & DMRA   & DANet & A2dele & Ours \\
\hline 
Model Size(MB)   & 951.9 & 929.7 & 533.6 &519 & 278 & 238.8 & 106.7 & 57.3 & 57.9/78.2   \\
\hline

FPS   & 19 & - & 15 &- & 7 & 10 & 32 & 120 & 136   \\
\hline

\end{tabular}
}
\caption{The model size and inference speed of different methods.}
\label{tab:model size}
\end{table*}

\section{Experiments}

\begin{figure*}
 \minipage{0.23\textwidth}
   \centering
   \includegraphics[width=\linewidth]{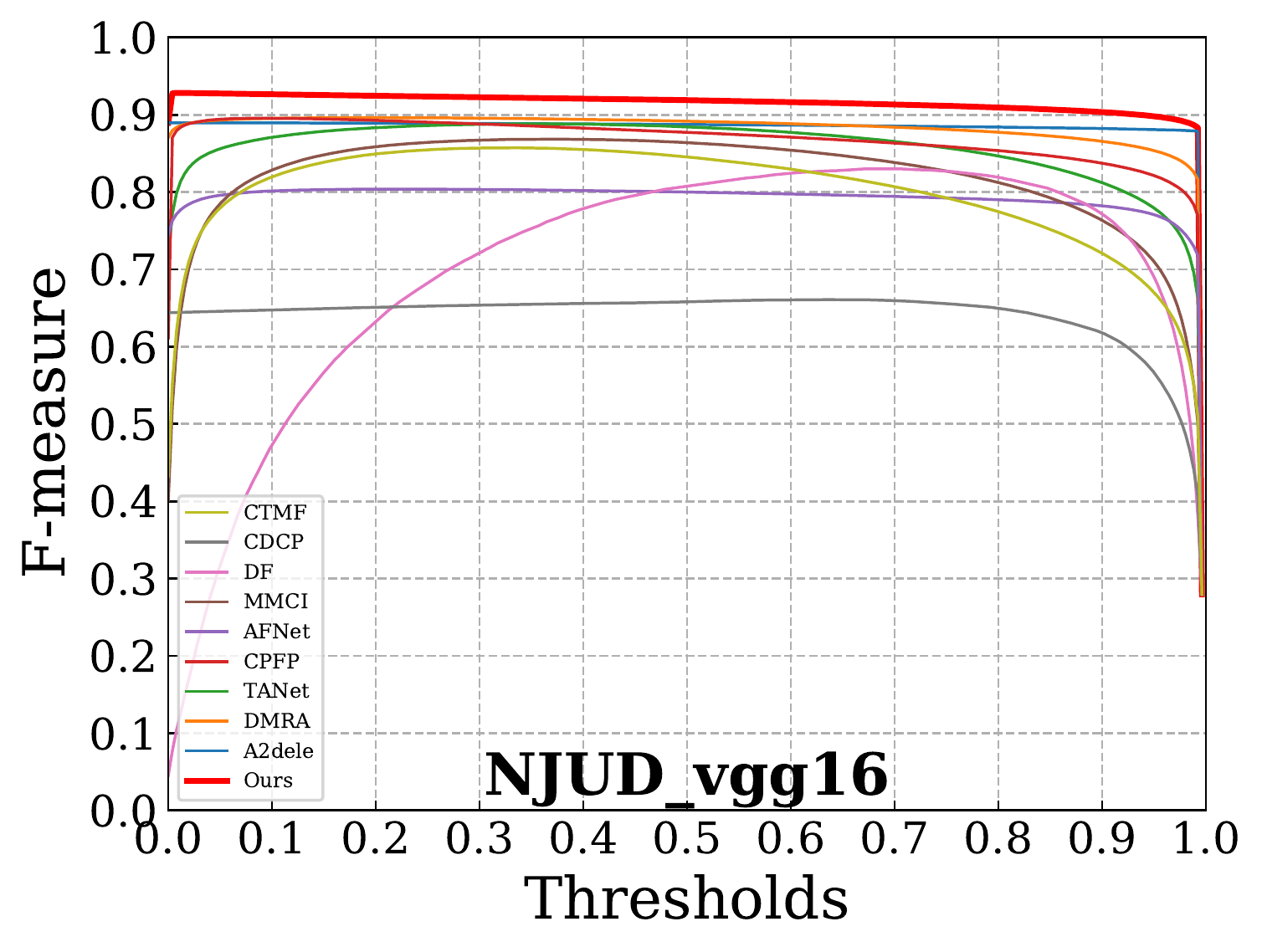}
   (a) 
 \endminipage\hfill
 \minipage{0.23\textwidth}%
   \centering
   \includegraphics[width=\linewidth]{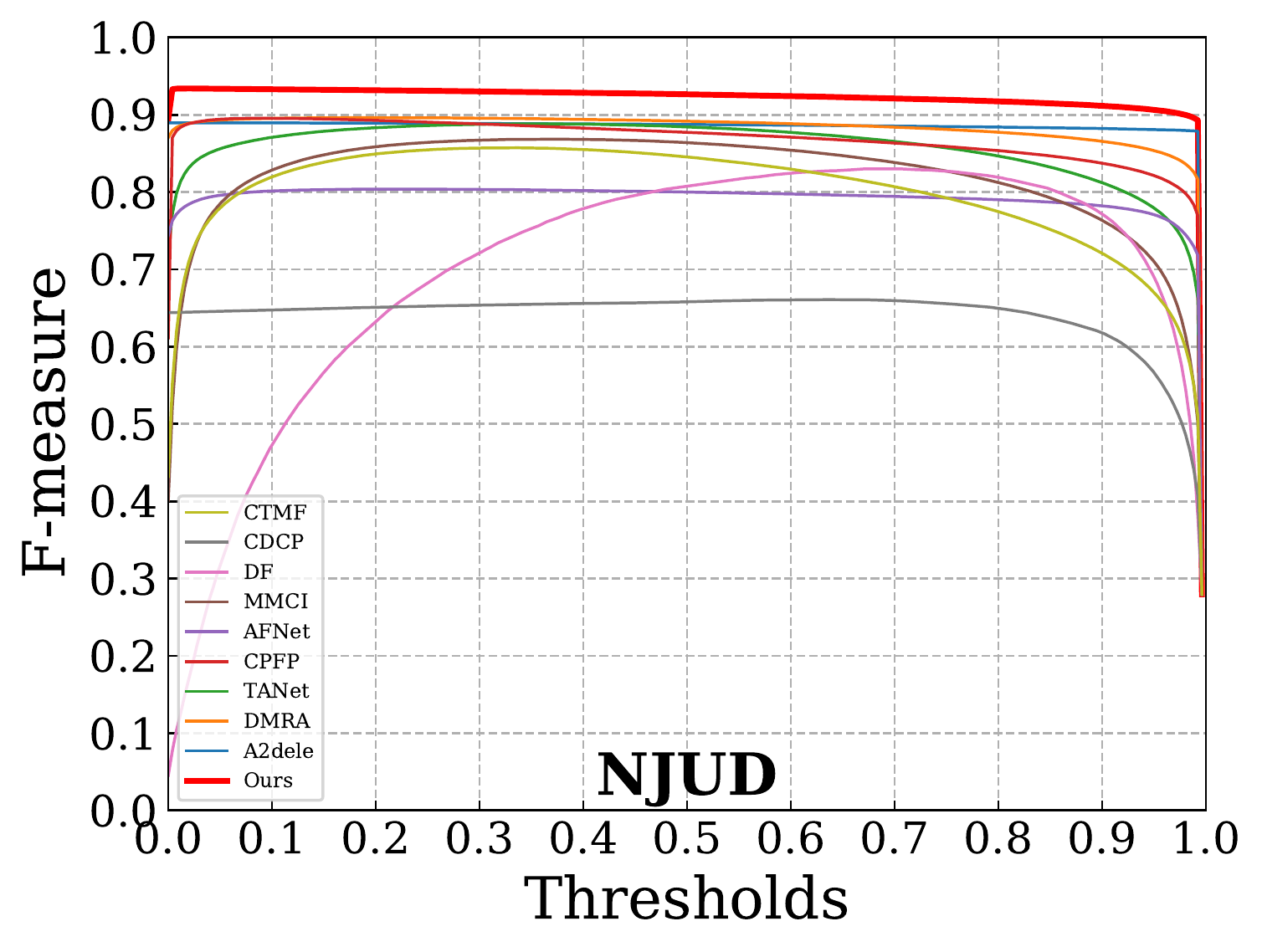}
   (b) 
 \endminipage\hfill
 \minipage{0.23\textwidth}
   \centering
   \includegraphics[width=\linewidth]{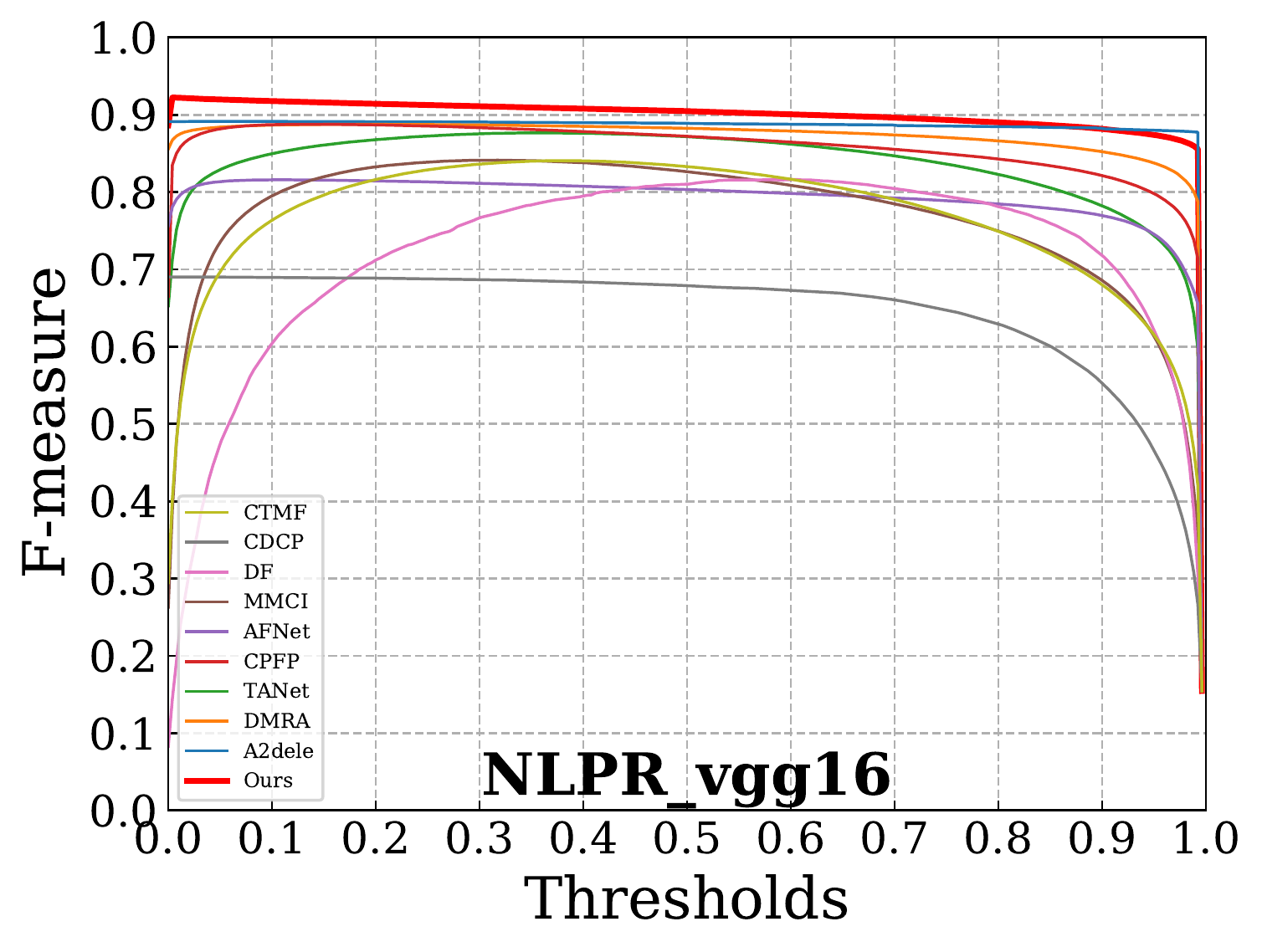}
   (c) 
 \endminipage\hfill
 \minipage{0.23\textwidth}%
   \centering
   \includegraphics[width=\linewidth]{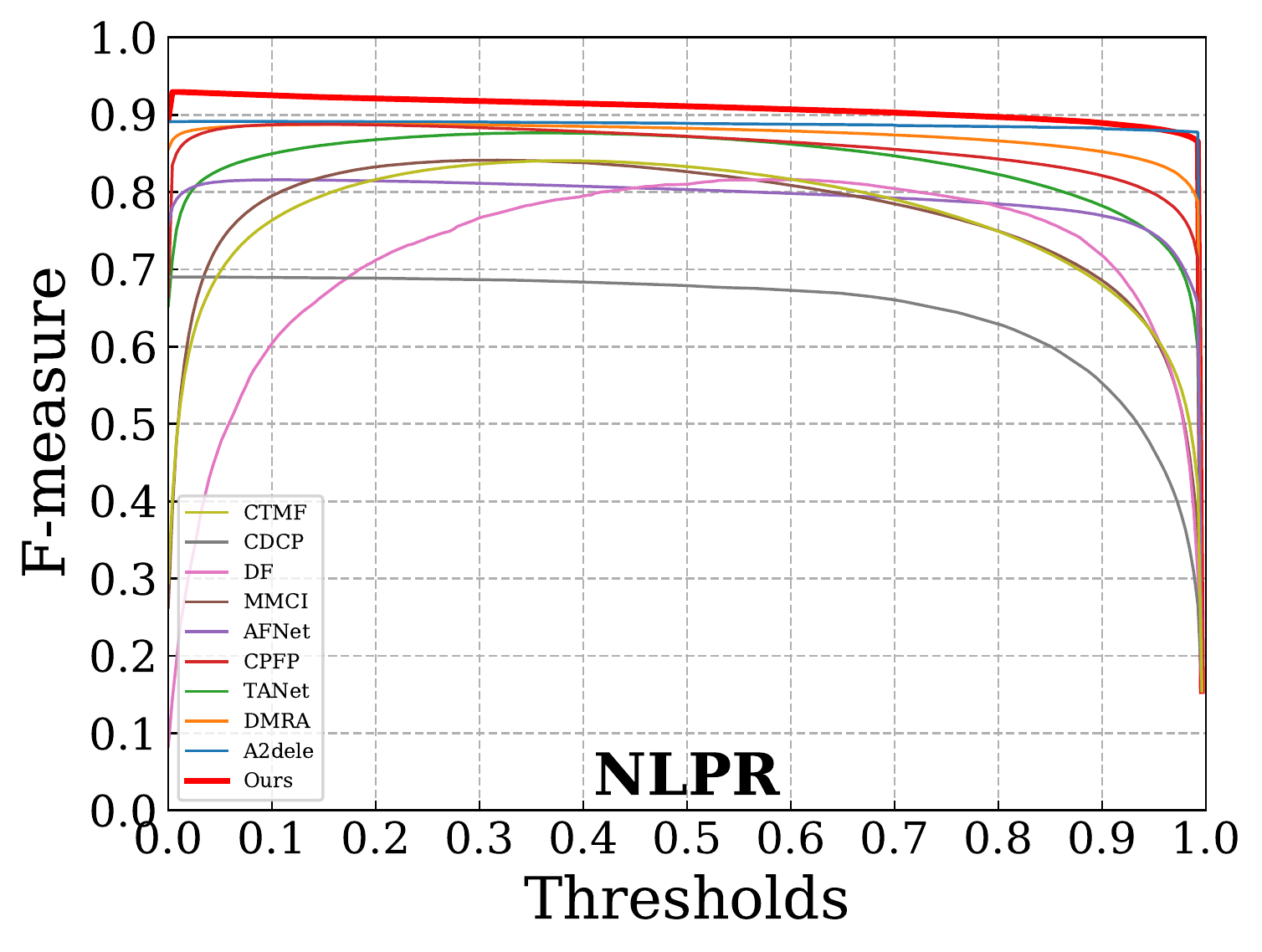}
   (d) 
 \endminipage\hfill
 \minipage{0.23\textwidth}%
  \centering
  \includegraphics[width=\linewidth]{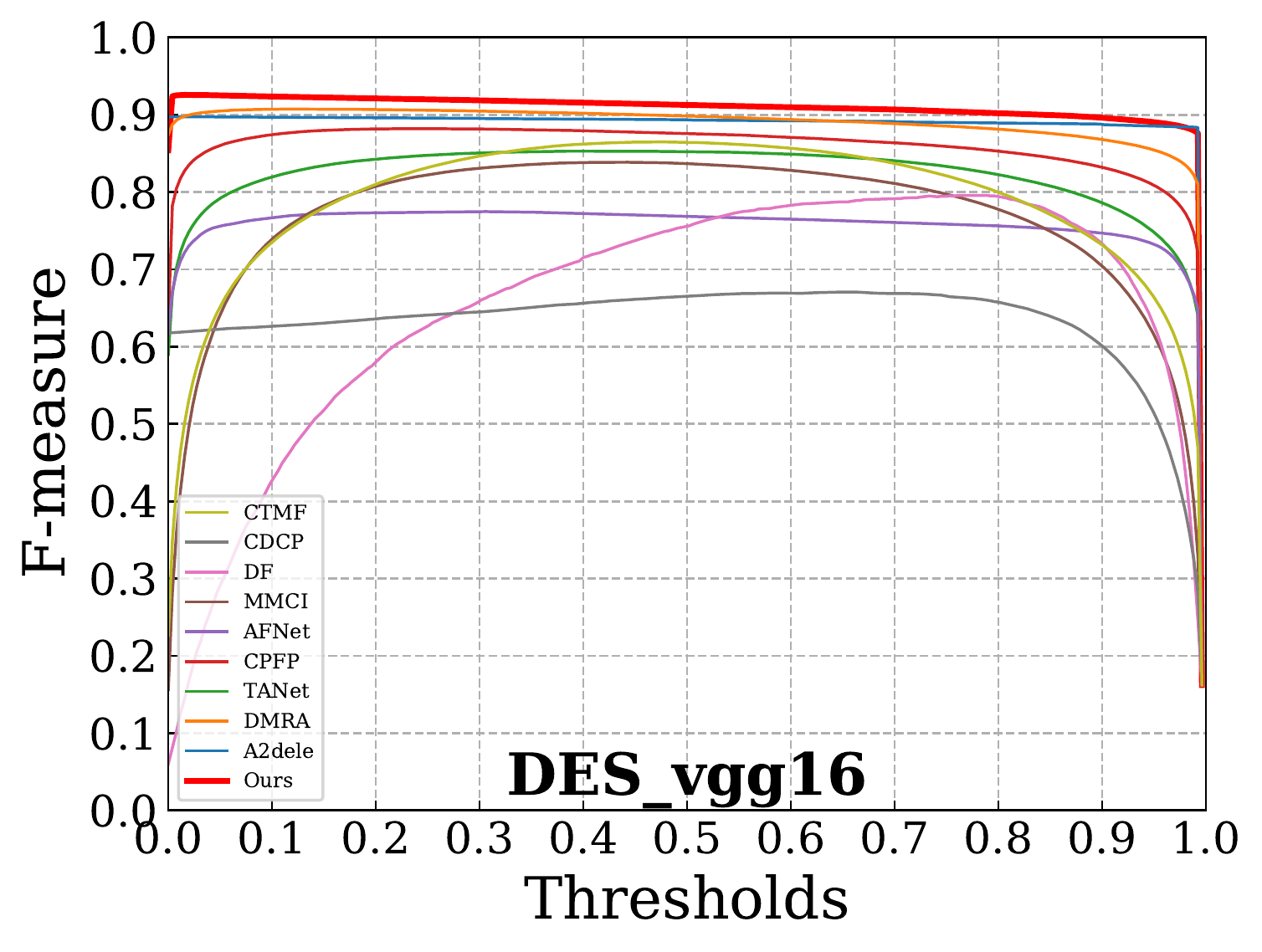}
  (e) 
 \endminipage\hfill
 \minipage{0.23\textwidth}%
  \centering
  \includegraphics[width=\linewidth]{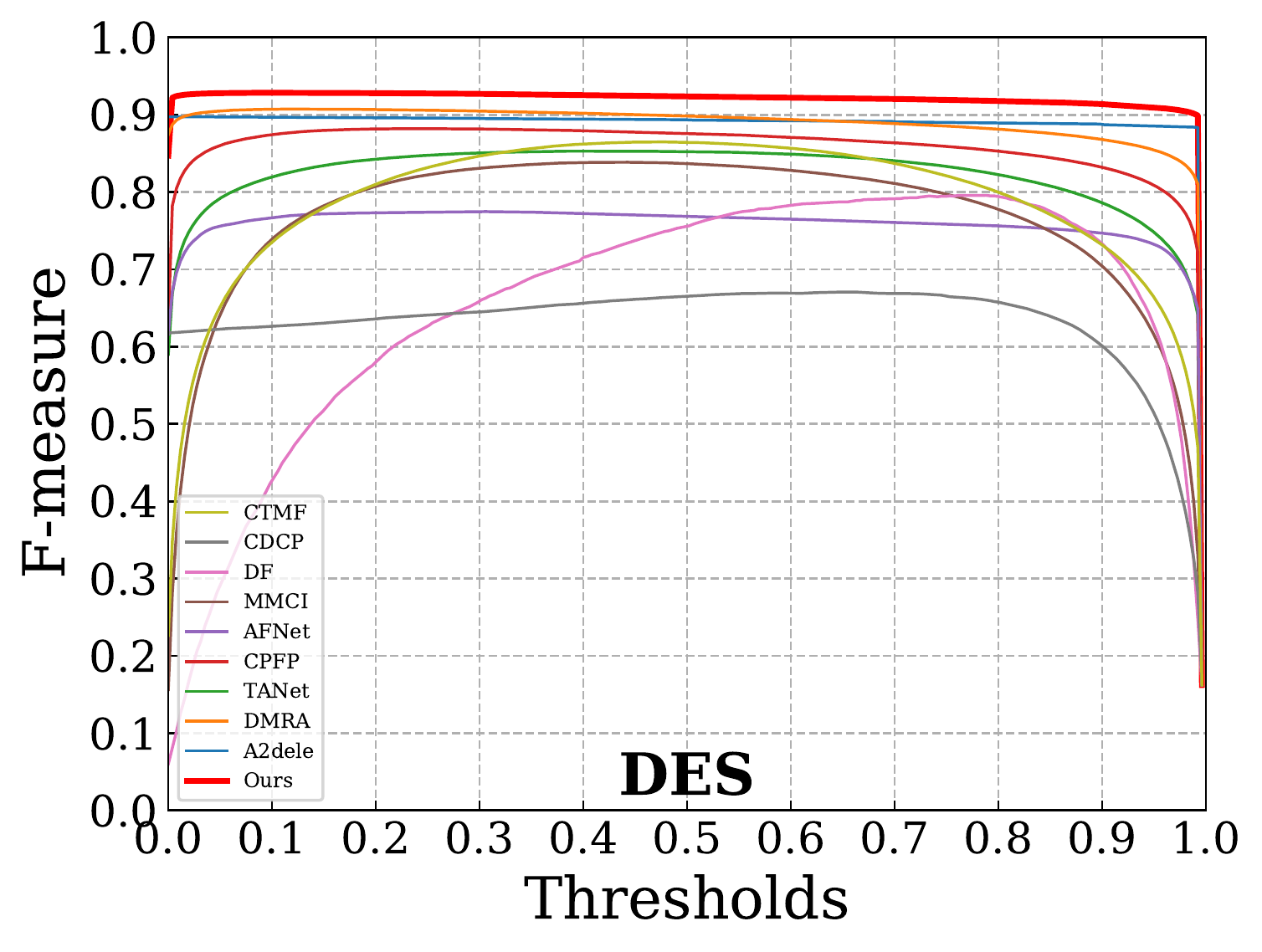}
  (f)
 \endminipage\hfill
 \minipage{0.23\textwidth}%
  \centering
  \includegraphics[width=\linewidth]{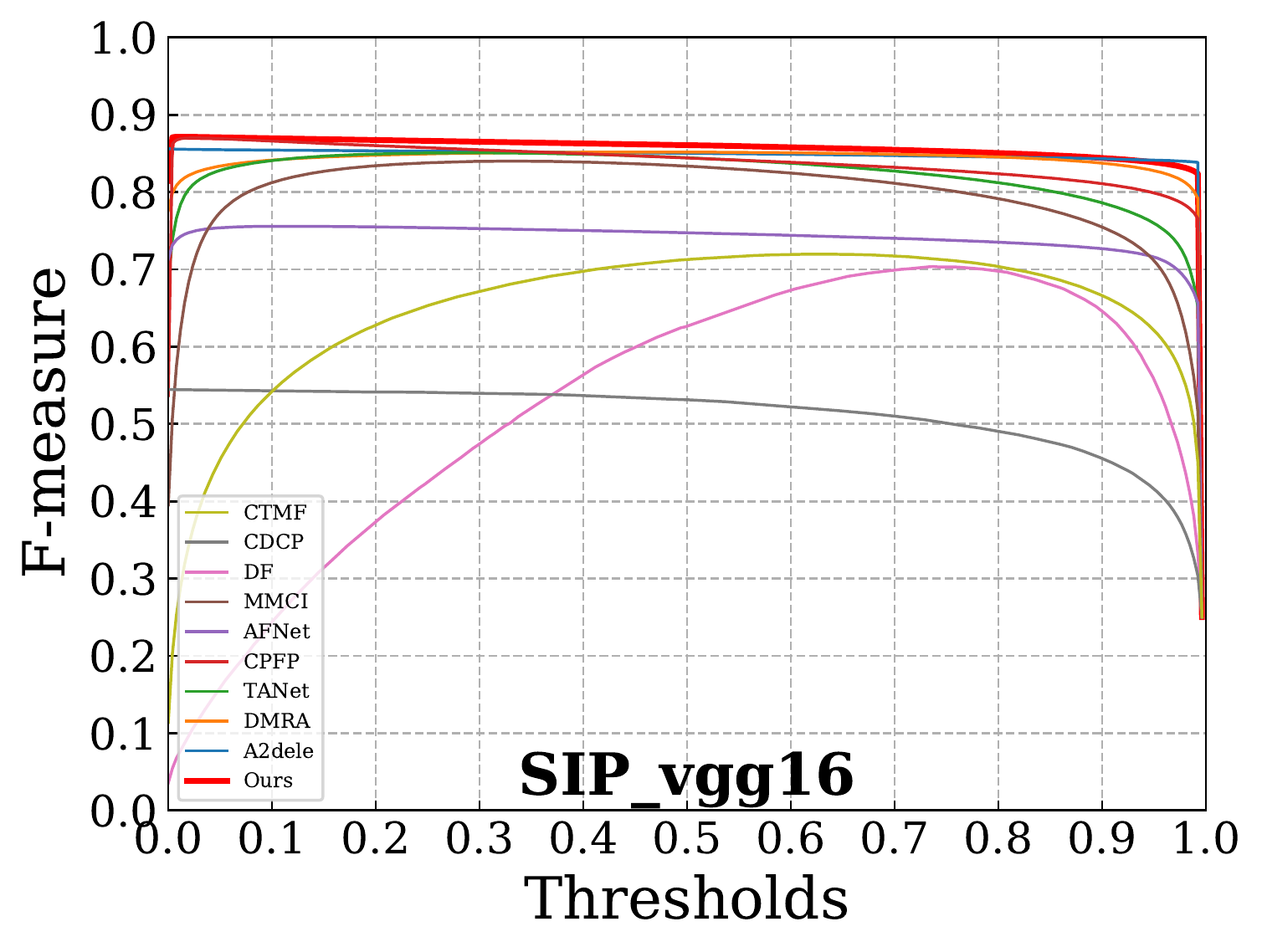}
  (g) 
 \endminipage\hfill
 \minipage{0.23\textwidth}%
  \centering
  \includegraphics[width=\linewidth]{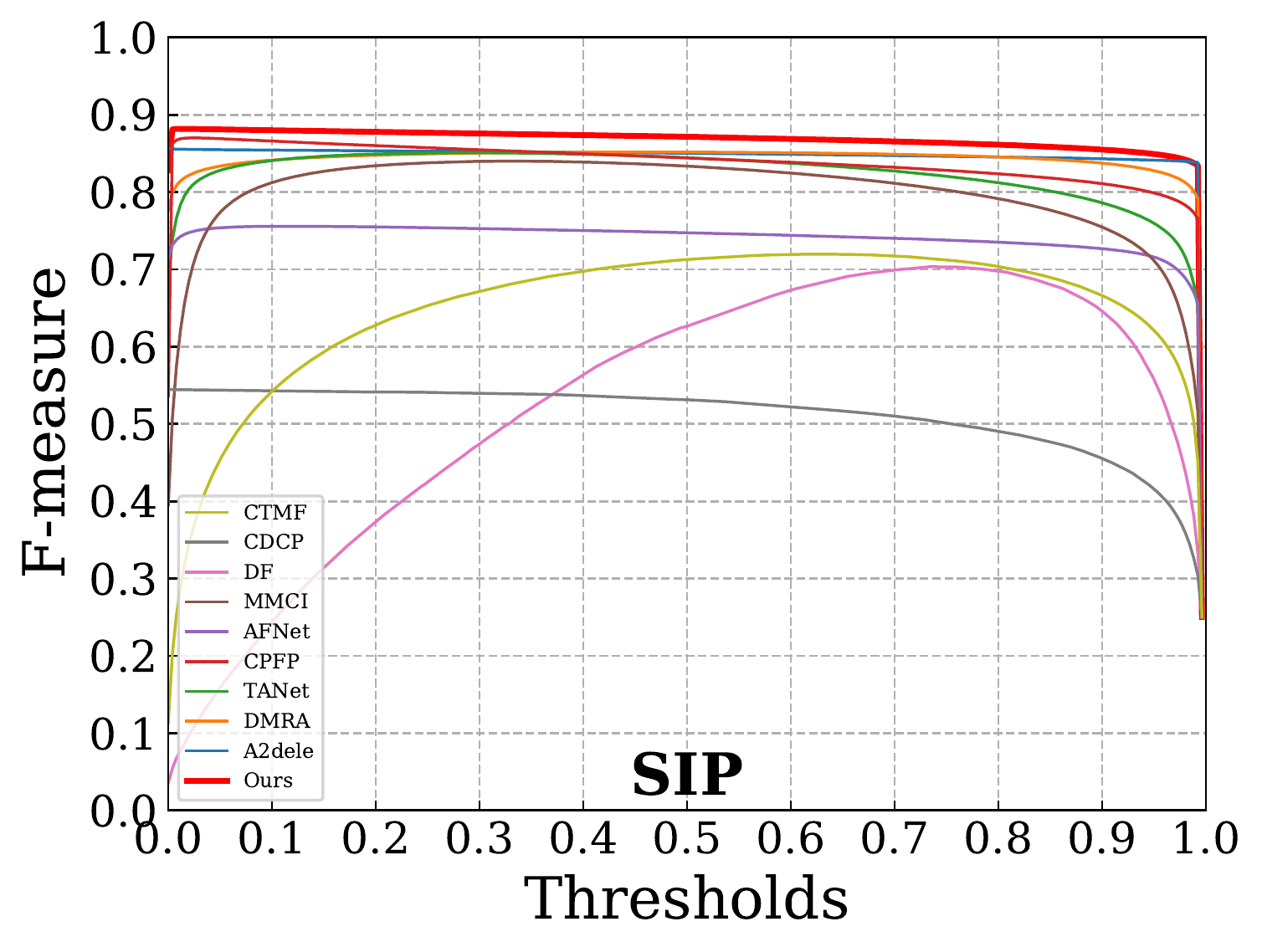}
  (h)
 \endminipage\hfill

\caption{Fm curves across four benchmarks on two different backbones.}
 \label{fig:FM curves for vgg16 and 19}
 \end{figure*}

\begin{figure*}[h]
%\minipage{0.5\textwidth}
  \centering
  \includegraphics[width=0.9\linewidth]{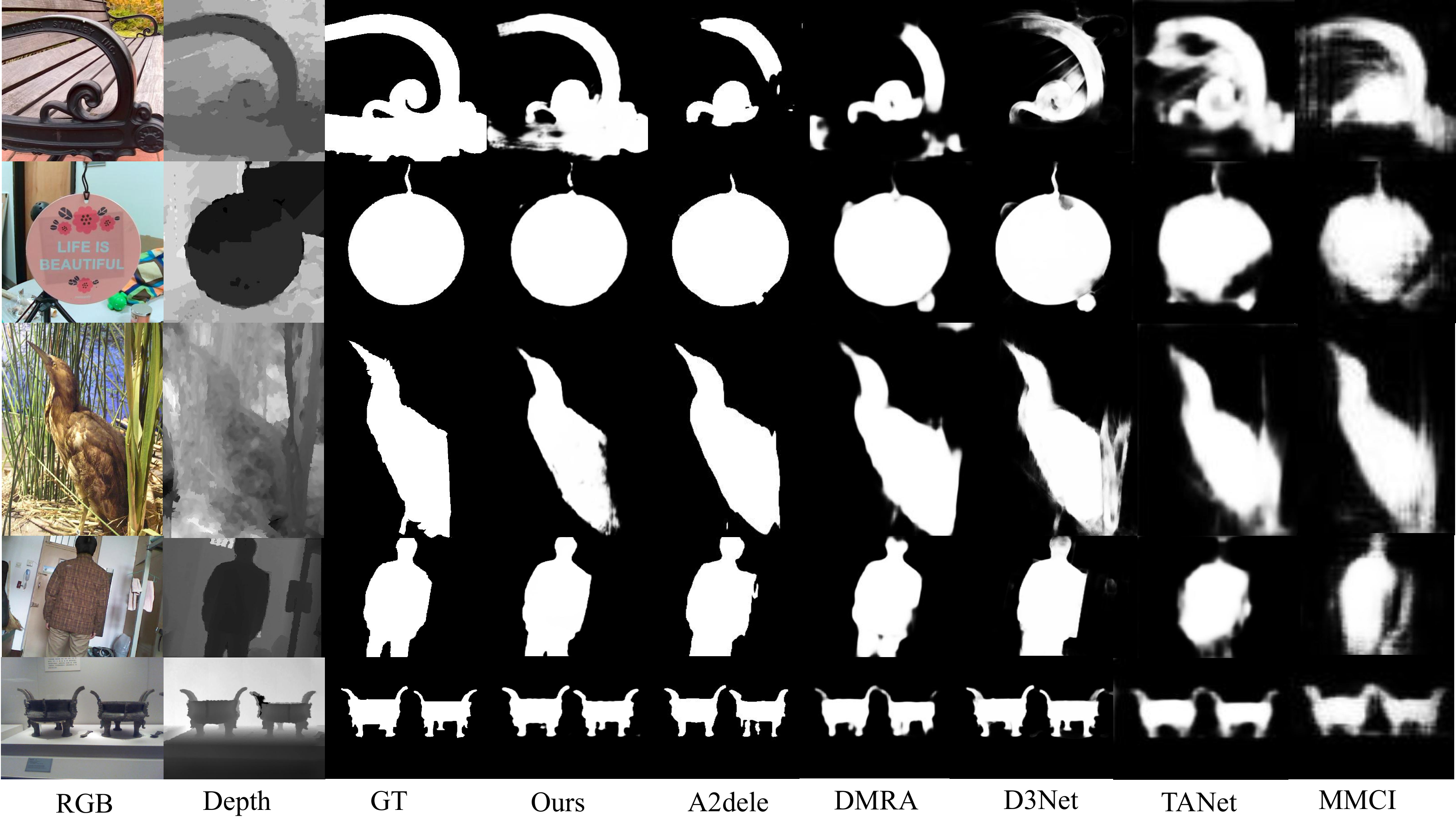}
  %(a) 
%\endminipage\hfill
\caption{Visual comparisons with existing methods}
\label{fig:visual comparisons of KD with sota}
\end{figure*}

\subsection{Datasets and Evaluation Metrics}
\noindent\textbf{Datasets.} Extensive experiments are conducted on five widely used RGB-D datasets, namely, NLPR~\cite{peng2014nlpr}, NJUD~\cite{ju2015njud}, SIP~\cite{DBLP:journals/D3/abs-1907-06781}, DES~\cite{cheng2014des} and LFSD~\cite{li2014lfsd}. These datasets contain large-scale images with different resolutions and diverse scenarios. We adopt the same training dataset with \cite{zhao2019contrast}, which contains 1500 samples from NJUD and 700 samples from NLPR. The rest images in these two datasets together with other three datasets are used for testing.
\\

\noindent\textbf{Evaluation Metrics.} We adopt five metrics to comprehensively evaluate SOD tasks. These metrics include the $F$-measure curves, the $F$-measure score ($F_{\beta}$), the Mean Absolute Error ($M$), the $S$-measure ($S_{\alpha}$) and the $E$-measure ($E_{\theta}$). Specifically,
$F_{\beta}$ measures the accuracy of the model as follows:

\begin{equation}
    \centering
    F_{\beta }=\dfrac {\left( 1+\beta ^{2}\right) \cdot \text{Precision}\cdot \text{Recall}}{\beta ^{2}\cdot \text{Precision}+\text{Recall}}
\end{equation}
where $\beta^{2}$ is set to 0.3 as default. $M$ measures the error rate of the model as follows:

\begin{equation}
    \centering
    M = \dfrac {1}{W\times H}\sum ^{W}_{x=1}\sum ^{H}_{y=1}\left| S\left( x,y\right) -G\left( x,y\right) \right|
\end{equation}
where $W$ denotes the width and $H$ denotes the height of prediction. $S$ is the prediction saliency map and $G$ is the corresponding ground truth.

\subsection{Implementation Details}
Our model is implemented using Pytorch Toolbox and trained on a GTX TITAN X GPU for 40 epochs with mini-batch size 4. We use a VGG16- and VGG19-based FPN as our final student architecture. Both RGB and depth images are resized to 256x256. To avoid overfitting, simple flipping and rotating are adopted to augment the training dataset. The initial learning rate is set to 1e-3 and we adopt a 0.0005 weight decay for the stochastic gradient descent (SGD) with a momentum of 0.9.

\begin{table*}[h]
\centering
\caption{Ablation analysis on 4 datasets. $RGB$ and $RGBD$ indicate that the student network is trained without/with depth maps respectively by cross-entropy loss. $s$ indicates the weight of knowledge distillation. Here we use the mean value of F-score $F_{m}$ and $wF_{m}$ to show the overall accuracy.}
\label{tab:distillation ablation in temperature 5}

\resizebox{0.8\linewidth}{!}{
\begin{tabular}{cc|ccccc|c|c}
\hline
\hline
\multirow{2}{*}{} & Metric &RGB  & RGBD & s=0.3  & s=0.5 & s=0.7 & s=Dynamic & +threshold \\
                        %  &  &  \cite{zhao2019contrast}  & \cite{DBLP:journals/D3/abs-1907-06781} &  \cite{zhao2020single}    & $*$  \\ \hline
                         \hline
\multirow{3}{*}{\rotatebox{90}{SIP}~\rotatebox{90}{}}    & $F_{m}\uparrow$ &0.704  & 0.773 & 0.832 & 0.845 &0.843  &0.849 & {\textcolor{black} {\textbf{0.853}}} \\
                     
                         & $wF_{m}\uparrow$  &0.654  &  0.724 & 0.796 & 0.805 &0.809 &0.811 & {\textcolor{black} {\textbf{0.817}}} \\
                         & $M\,\downarrow$ &0.108  & 0.086 & 0.063 & 0.061 &0.059 &0.058 & {\textcolor{black} {\textbf{0.056}}} \\ \hline

\multirow{3}{*}{\rotatebox{90}{NJUD}~\rotatebox{90}{}}    & $F_{m}\uparrow$ &0.776  & 0.830 & 0.902 & 0.902 &0.898  &0.904 & {\textcolor{black} {\textbf{0.914}}} \\
                     
                         & $wF_{m}\uparrow$  &0.739  &  0.799 & 0.895 & 0.889 &0.880 &0.893 & {\textcolor{black} {\textbf{0.901}}} \\
                         & $M\,\downarrow$ &0.080  & 0.060 & 0.030 & 0.034 &0.037 &0.032 & {\textcolor{black} {\textbf{0.030}}} \\ \hline

\multirow{3}{*}{\rotatebox{90}{NLPR}~\rotatebox{90}{}}    & $F_{m}\uparrow$ &0.780  & 0.816 & 0.873 & 0.876 &0.870  &0.876 & {\textcolor{black} {\textbf{0.890}}} \\
                     
                         & $wF_{m}\uparrow$  &0.746  &  0.781 & 0.877 & 0.875 &0.865 &0.876 & {\textcolor{black} {\textbf{0.887}}} \\
                         & $M\,\downarrow$ &0.046  & 0.041 & 0.022 & 0.024 &0.026 &0.024 & {\textcolor{black} {\textbf{0.021}}} \\ \hline

\multirow{3}{*}{\rotatebox{90}{LFSD}~\rotatebox{90}{}}    & $F_{m}\uparrow$ &0.713  & 0.784 & 0.825 & 0.832 &0.830  &0.834 & {\textcolor{black} {\textbf{0.835}}} \\
                     
                         & $wF_{m}\uparrow$  &0.656  &  0.741 & 0.780 & 0.793 &0.790 &0.795 & {\textcolor{black} {\textbf{0.796}}} \\
                         & $M\,\downarrow$ &0.142  & 0.102 & 0.086 & 0.080 &0.080 &0.078 & {\textcolor{black} {\textbf{0.078}}} \\ \hline

\hline

\end{tabular}%}
%\end{center}
}
\end{table*}

\subsection{Comparisons with the state of arts}
We conduct our experiments on 10 different prevalent methods in recent years, including DF~\cite{qu2017df} , CTMF~\cite{han2017ctmf}, AFNet~\cite{wang2019afnet}, MMCI~\cite{chen2019mmci}, DMRA~\cite{piao2019dmra}, CPFP~\cite{zhao2019contrast}, TANet~\cite{chen2019tanet}, D3Net~\cite{DBLP:journals/D3/abs-1907-06781}, A2delde~\cite{piao2020a2dele}  and DANet~\cite{zhao2020danet}. For fair comparisons, we directly use the released evaluation results or generate the results by the public saliency maps under the same evaluation framework.
\\

\begin{figure*}[h!]
\minipage{0.2\textwidth}
  \centering
  \includegraphics[width=\linewidth]{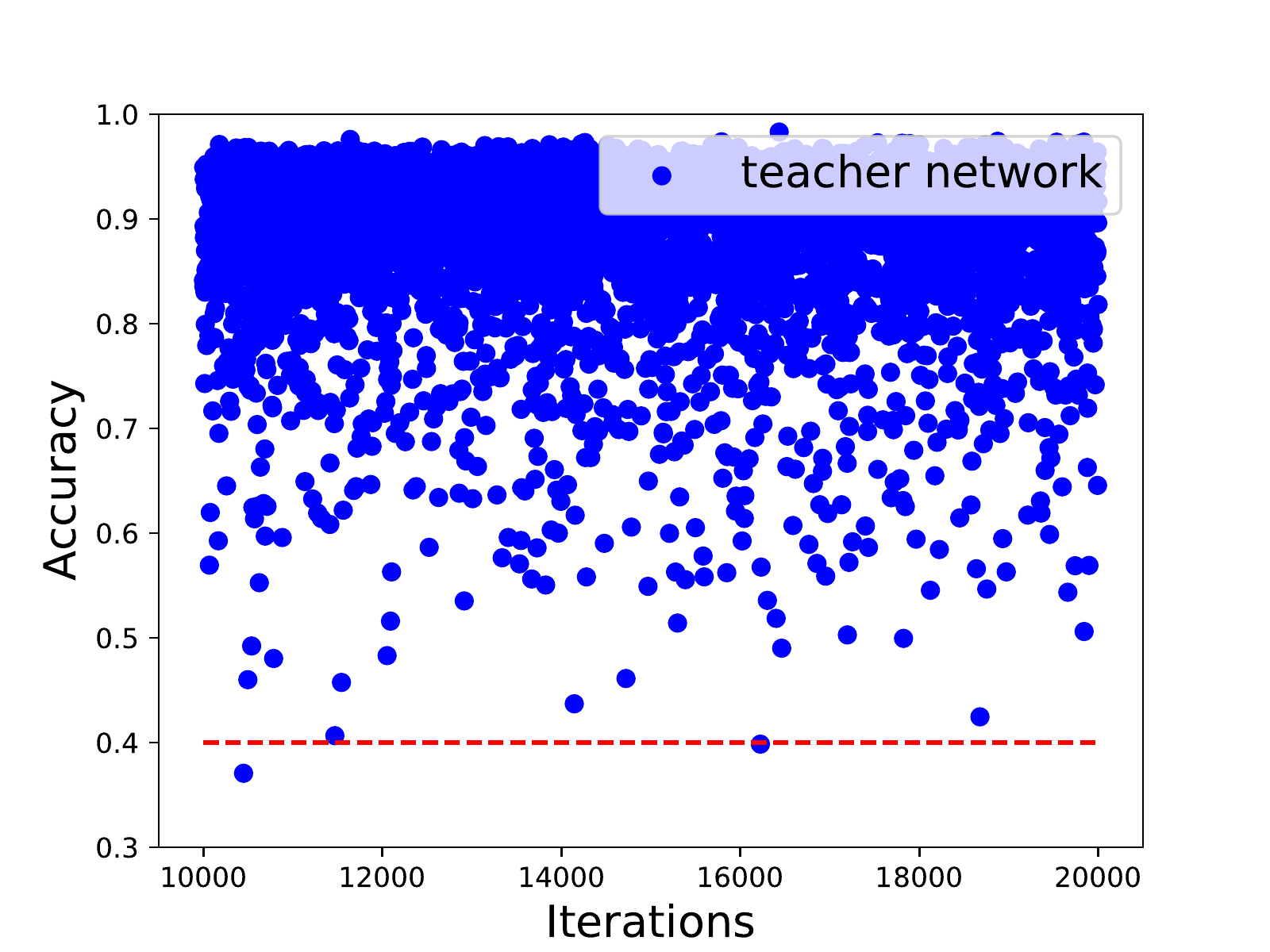}
  (a) teacher
\endminipage\hfill
\minipage{0.2\textwidth}%
  \centering
  \includegraphics[width=\linewidth]{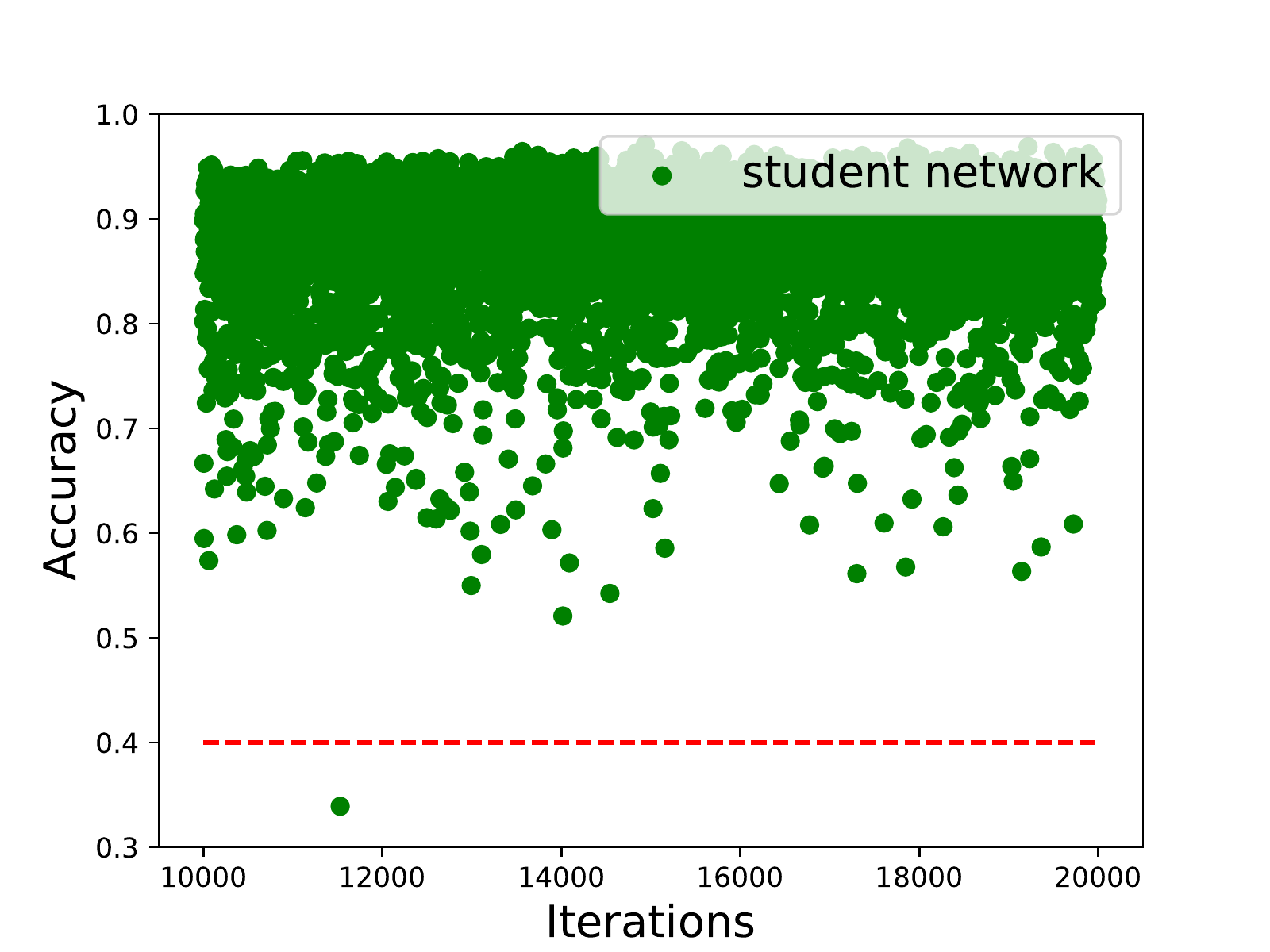}
  (b) 0.3
\endminipage\hfill
\minipage{0.2\textwidth}%
  \centering
  \includegraphics[width=\linewidth]{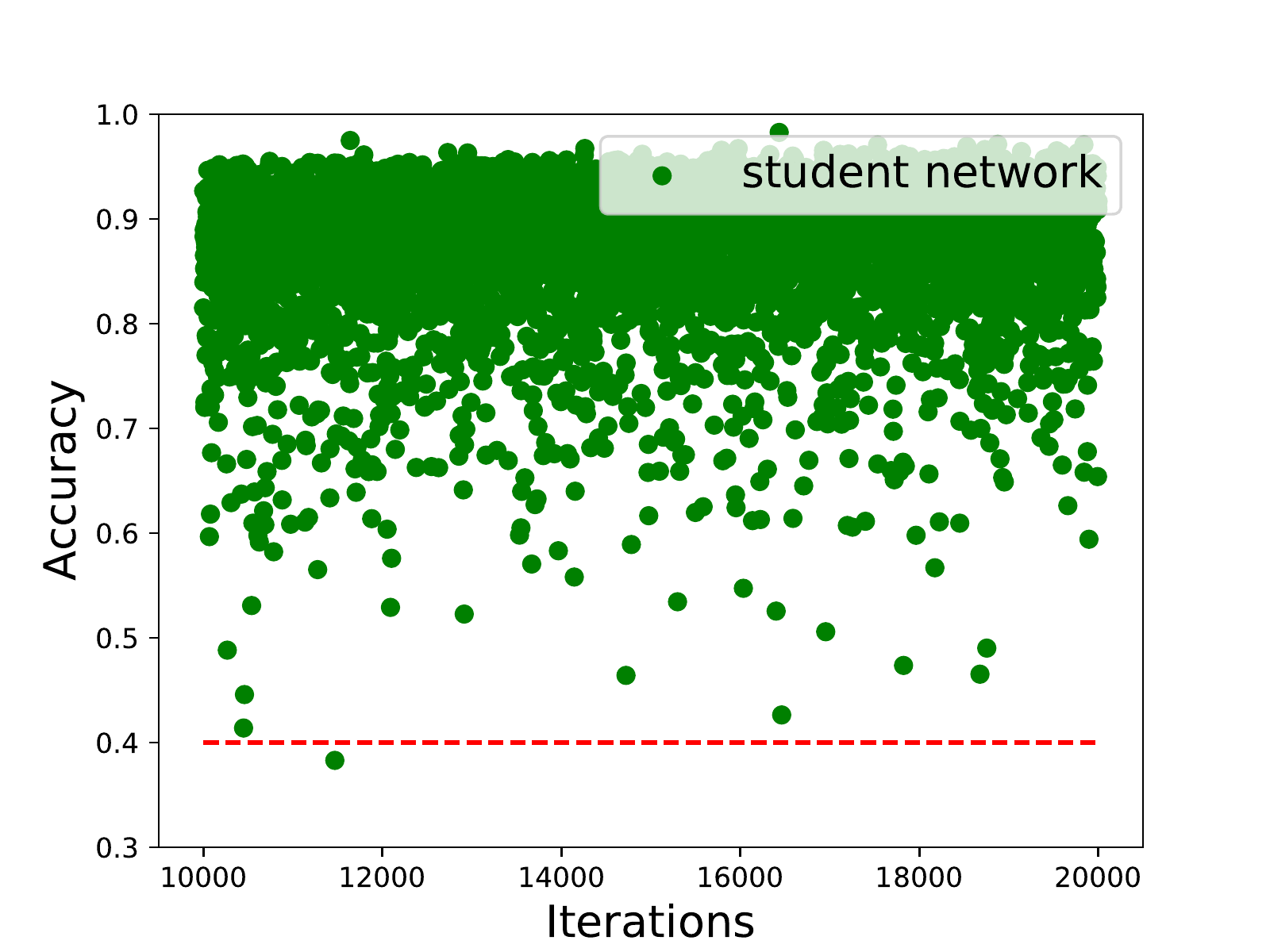}
  (c) 0.5
\endminipage\hfill
\minipage{0.2\textwidth}%
  \centering
  \includegraphics[width=\linewidth]{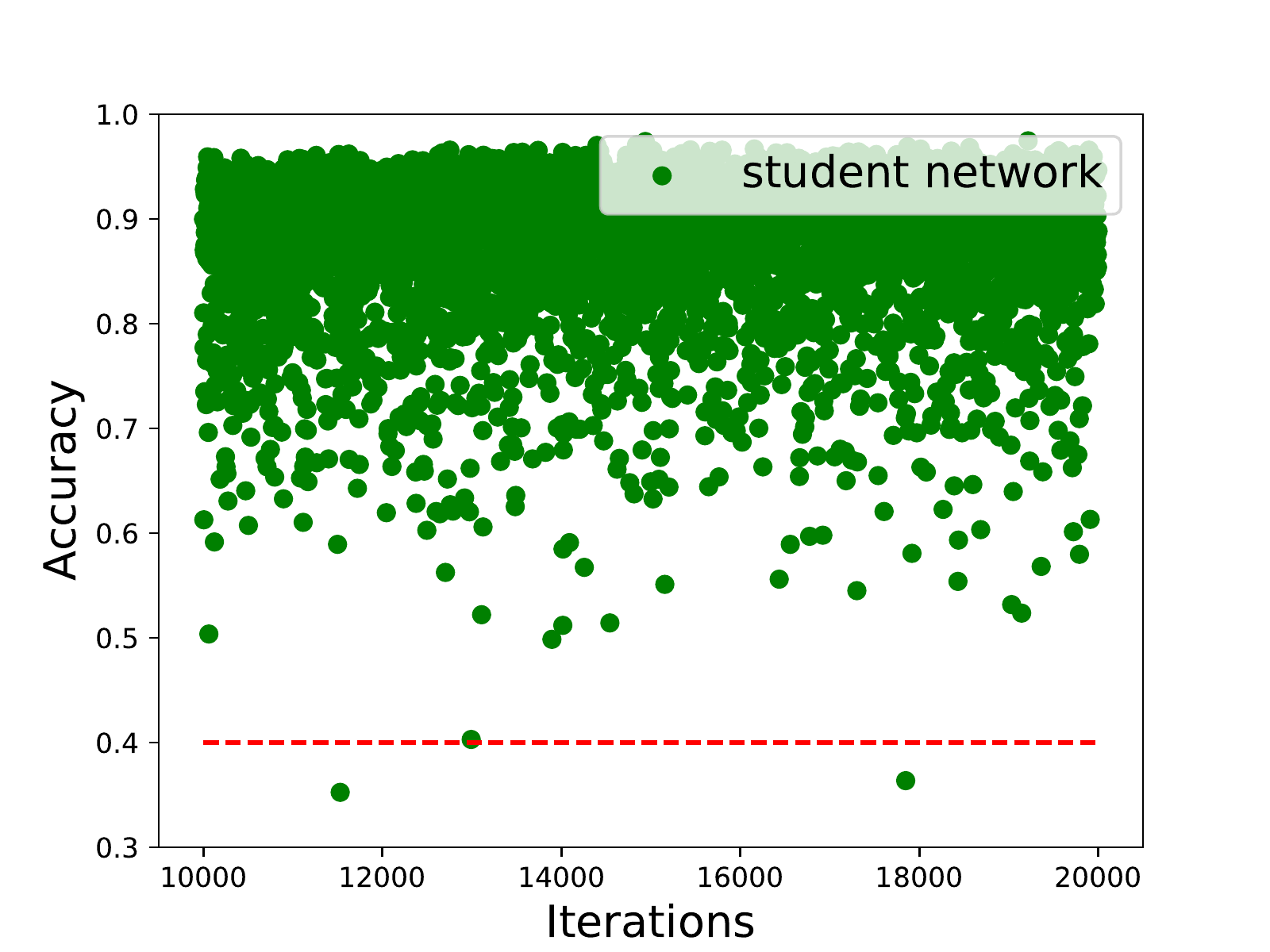}
  (d) 0.7
\endminipage\hfill
\minipage{0.2\textwidth}%
  \centering
  \includegraphics[width=\linewidth]{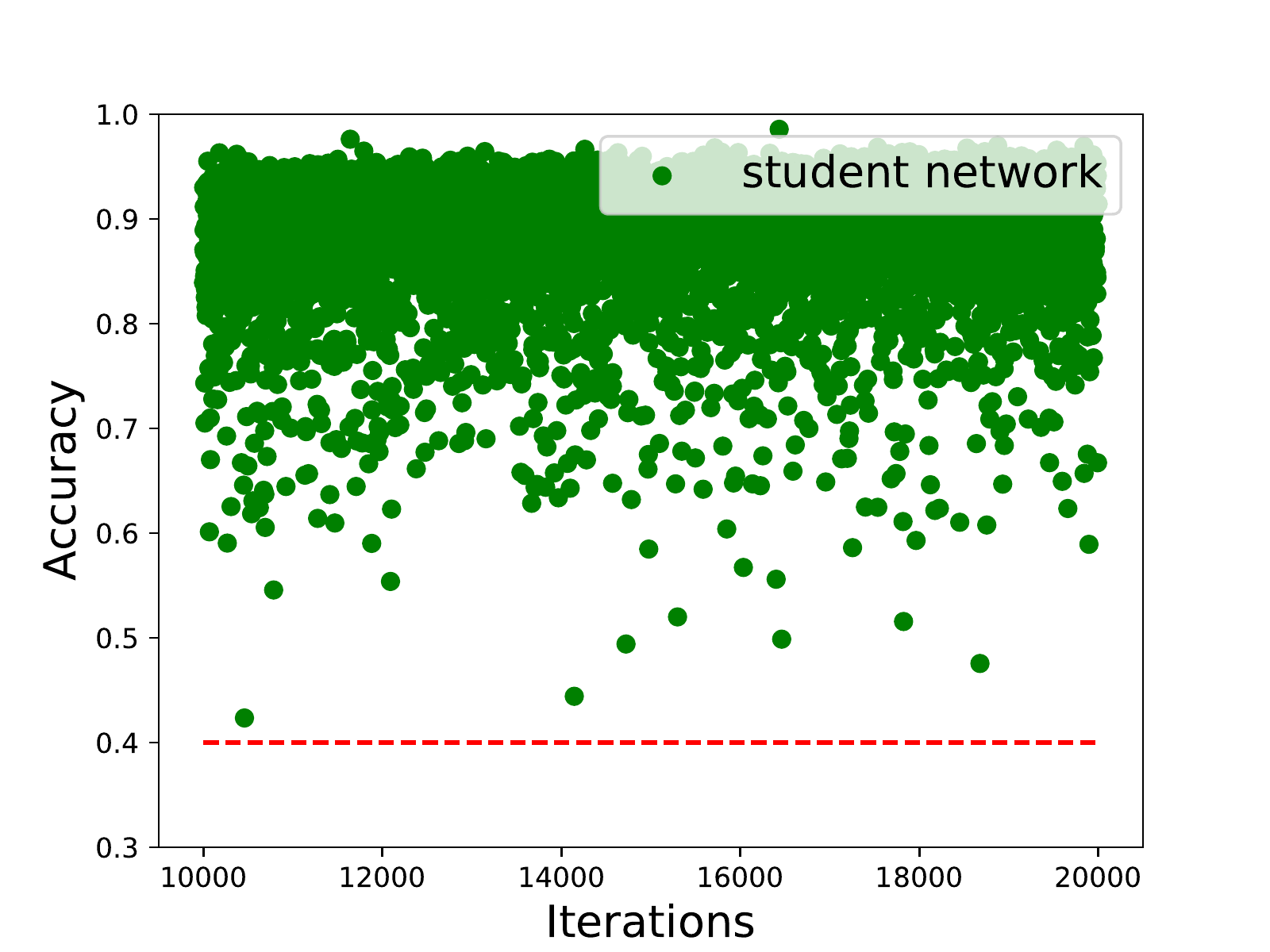}
  (e) dynamic
\endminipage\hfill
\caption{Teacher and student scatters with different distillation weights. We exhibit the last 10000 training iterations and dynamic KD shows better overall performance.}
\label{fig:scatters of different distillation weights}
\end{figure*}

\begin{figure}
%\minipage{0.8\textwidth}
  \centering
  \includegraphics[width=\linewidth]{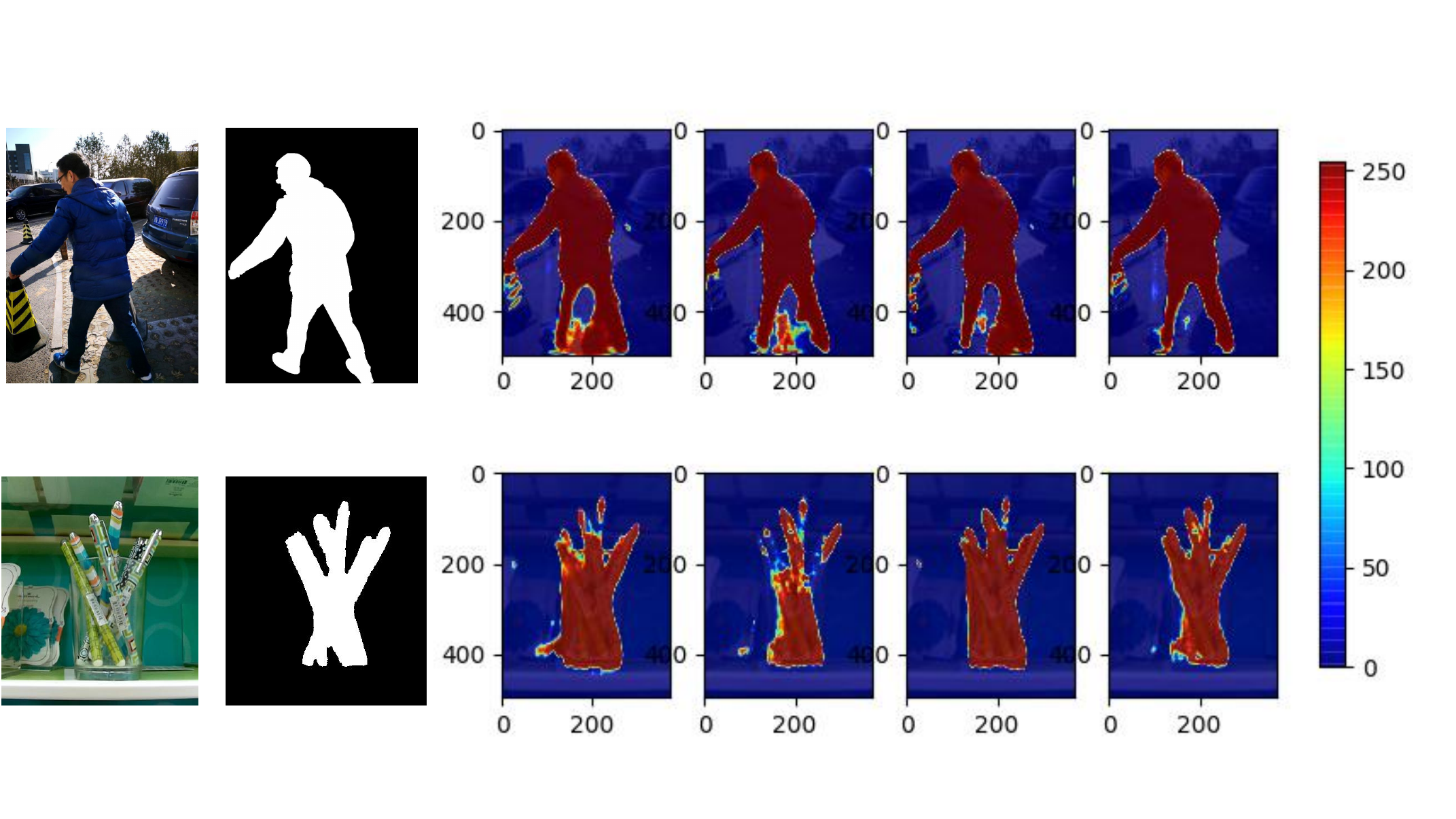}

\caption{Heatmap visual comparisons of different distillation weights: heatmaps from left to right indicate weights 0.3, 0.5, 0.7 and dynamic respectively.}
\label{fig:dynamic distill heatmap}
\end{figure}

\noindent\textbf{Quantitative Evaluation.} Table \ref{table:overall quantitive in distillation} shows the quantitative results over five datasets. It can be observed that our method achieves the best scores in most metrics, especially on the NJUD dataset which contains 500 testing image pairs, for which our method performs better as far as all metrics are concerned. As for LFSD and SIP, although higher results come from other methods, we still obtain competing results in smaller VGG16 and VGG19 based networks. Figure \ref{fig:FM curves for vgg16 and 19} shows the comparison results using one-dimensional curves. Our method is represented by the red line which demonstrates better overall performance in both lightweight student models. Besides, it is apparent in Table \ref{tab:model size} that our final VGG16-based network only has 57.9MB with a faster speed, which drastically improves the inference speed and reduces the number of parameters. The above results indicate that without designing complicated models, accurate detection results can be obtained by only using an FPN with the help of the proposed methods.
\\

\noindent\textbf{Qualitative Evaluation.}
Figure \ref{fig:visual comparisons of KD with sota} exhibits the visual comparisons with prevalent methods in recent years. Images contain diverse objects and scenarios, which are picked from different testing datasets. It can be observed that the saliency maps generated by our method are closer to the ground truth. More specifically, row 1 shows the case where the depth image has low contrast especially on the bottom part and row 2 and 3 show complex backgrounds in the RGB images. Under these circumstances, our method generates better saliency maps with less distortion and irrelevant objects compared to other methods.

\subsection{Ablation Studies}

\noindent\textbf{Dynamic Knowledge Distillation.}
As shown in Table \ref{tab:distillation ablation in temperature 5}, our baseline is an FPN with VGG19 backbone trained on a cross-entropy loss, which can achieve the basic detection task. $RGB$ indicates that only using RGB maps in training and $RGBD$ concatenates depth maps as input. It is worth noting that directly using early fusion strategy also shows potential in RGB-D saliency detection.
Then we employ KD on the baseline to compare the results in different weights on four datasets. It is observed that KD can improve the performance and our dynamic KD achieves better results across four testing datasets.
% especially on $M$, which indicates that our method can learn more reliable and task-relevant information from the teacher network. The visual comparisons in Figure \ref{fig:ablations}(1) illustrate the same conclusion through the saliency maps and heatmaps. 
Furthermore, Figure \ref{fig:scatters of different distillation weights} shows detection performance of the student network in different KD weights. It is observed that in the last 10000 iterations of training stage, the proposed method has better overall accuracy, where the lowest accuracy is still above 0.4, which is even better than the teacher network. Specific examples in Figure \ref{fig:dynamic distill heatmap} illustrates that compared with the dynamic KD, conventional fixed weights suffer from more false positives and true negatives. Consequently, it is demonstrated that our dynamic KD adaptively controls the knowledge distillation in an appropriate way, leading to the improvement of overall detection performance.

\begin{figure}[h]
%\minipage{0.5\textwidth}
  \centering
  \includegraphics[width=\linewidth]{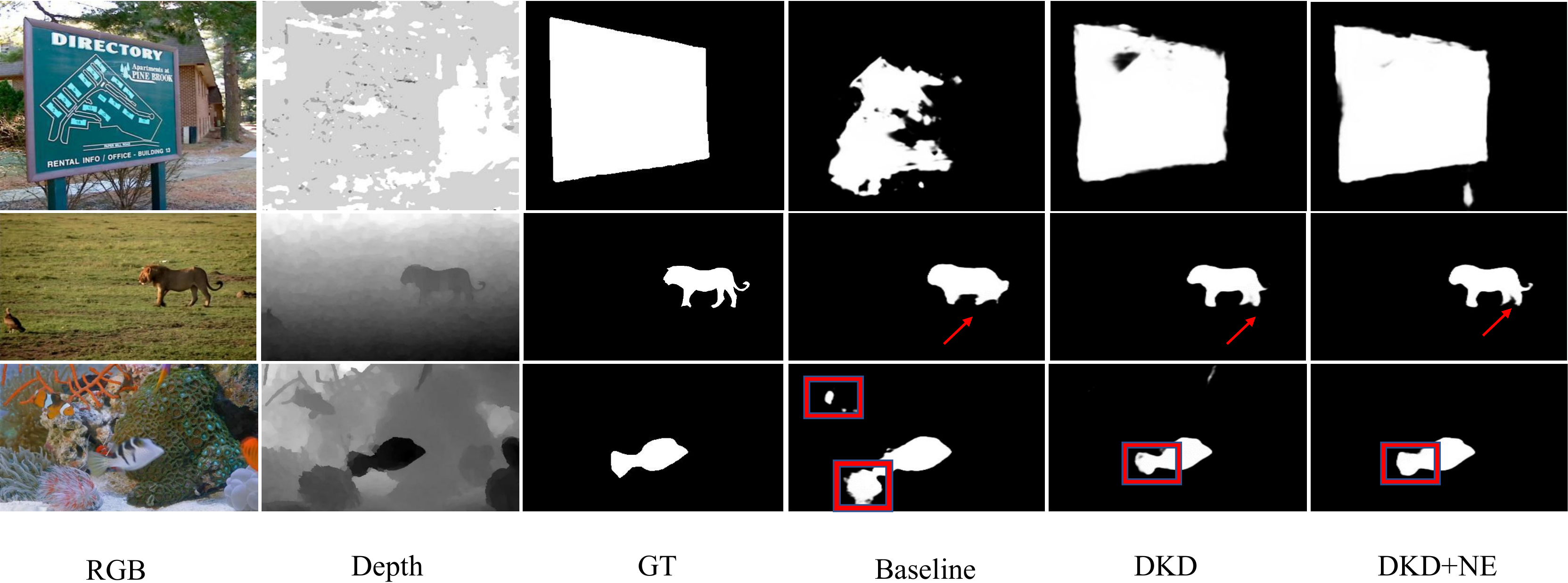}
  %(a) 
%\endminipage\hfill
\caption{Ablation studies of the proposed methods. $Baseline$ represents that the student network is only trained on cross-entropy loss. $DKD$ represents the proposed dynamic knowledge distillation and $NE$ means noise elimination.}
\label{fig:visual ablation study of KD}
\end{figure}

\noindent\textbf{Noise elimination with dynamic knowledge distillation.}
We investigate the low accuracy issue in teacher network by visualising extremely hard samples which has been shown in Figure \ref{fig:low quality depth}. It is observed in Table \ref{tab:distillation ablation in temperature 5} that with the help of noise elimination, all evaluation metrics over four testing datasets approach better results. Red arrows and rectangles in Figure\ref{fig:visual ablation study of KD} label the details which are refined by the proposed methods, especially on the part of object in the low contrast background, further illustrating that the proposed noise elimination effectively mitigate the noise during distillation and make the student model be able to learn more semantic details in the useful training data.

\begin{table}[h]
\centering
\caption{Quantitative comparisons of applying DKD on DANet.}
\label{tab:DANet distillation on VGG19}

\resizebox{0.8\linewidth}{!}{
\begin{tabular}{cc|cc}
\hline
\hline
\multirow{2}{*}{} & Metric    & DANet   & DANet+DKD  \\
                        %  &  &  \cite{zhao2019contrast}  & \cite{DBLP:journals/D3/abs-1907-06781} &  \cite{zhao2020single}    & $*$  \\ \hline
                         \hline

\multirow{3}{*}{\rotatebox{90}{SIP}~\rotatebox{90}{}}    & $F_{m}\uparrow$  & 0.864 & 0.848  \\
                     
                         & $wF_{m}\uparrow$   & 0.829 & 0.811  \\
                         & $M\,\downarrow$  & 0.054 & 0.058 \\ \hline

\multirow{3}{*}{\rotatebox{90}{DES}~\rotatebox{90}{}}    & $F_{m}\uparrow$  & 0.891 & 0.892  \\
                     
                         & $wF_{m}\uparrow$   & 0.848 & 0.870  \\
                         & $M\,\downarrow$  & 0.028 & 0.025  \\ \hline
                         
\multirow{2}{*}{}    & Model Size(MB)  & 106.7 & 78.2  \\
 \hline

\hline

\end{tabular}%}
%\end{center}
}
\end{table}

.

\begin{table}[h]
\centering
\caption{Distillation on VGG16 with RGB maps.}
\label{tab:distillation ablation only on RGB with temperature 10}

\resizebox{0.8\linewidth}{!}{
\begin{tabular}{cc|ccc|c}
\hline
\hline
\multirow{2}{*}{} & Metric    & s=0.3  & s=0.5 & s=0.7 & s=Dynamic  \\
                        %  &  &  \cite{zhao2019contrast}  & \cite{DBLP:journals/D3/abs-1907-06781} &  \cite{zhao2020single}    & $*$  \\ \hline
                         \hline

\multirow{3}{*}{\rotatebox{90}{NLPR}~\rotatebox{90}{}}    & $F_{m}\uparrow$  & 0.882 & 0.876 &0.876   & {\textcolor{black} {\textbf{0.884}}} \\
                     
                         & $wF_{m}\uparrow$   & 0.869 & 0.865 &0.866  & {\textcolor{black} {\textbf{0.879}}} \\
                         & $M\,\downarrow$  & 0.024 & 0.025 &0.027  & {\textcolor{black} {\textbf{0.024}}} \\ \hline

\multirow{3}{*}{\rotatebox{90}{LFSD}~\rotatebox{90}{}}    & $F_{m}\uparrow$  & 0.790 & 0.786 &0.788   & {\textcolor{black} {\textbf{0.798}}} \\
                     
                         & $wF_{m}\uparrow$   & 0.745 & 0.735 &0.751  & {\textcolor{black} {\textbf{0.753}}} \\
                         & $M\,\downarrow$  & 0.104 & 0.106 &0.101  & {\textcolor{black} {\textbf{0.097}}} \\ \hline

\hline

\end{tabular}%}
%\end{center}
}
\end{table}

\noindent\textbf{Further analysis.} In order to show the generalization of the proposed DKD, we replace the teacher network to DANet. Experimental results in Table \ref{tab:DANet distillation on VGG19} indicate that our DKD can compress the model size of existing method and approach similar accuracy to the teacher model.
We further conduct experiments on VGG16-based FPN with different KD hyper-parameters as shown in Table \ref{tab:distillation ablation only on RGB with temperature 10}. Specifically, we set temperature to 10 and only use the RGB images in the distillation training phase. Experimental results demonstrate that the proposed DKD can be utilized on different networks with different training settings, proving the effectiveness and generalization of DKD and leading to the potential of achieving RGB-D SOD tasks through RGB data within a lightweight structure.

\section{Conclusions}
In this paper, we propose a dynamic knowledge distillation strategy and a noise elimination method for RGB-D based SOD. The proposed dynamic strategy considers the performance of both teacher and student models to generate an adaptive weight for knowledge distillation. Besides, we investigate the noise issue caused by depth maps and alleviate this problem by setting a threshold in KD. The propose methods provide a new perspective which avoids designing extra networks for RGB-D SOD. We conduct comprehensive experiments on five challenging benchmark datasets to demonstrate that our method achieves competitive performance by only using a simple FPN model, which significantly compresses the model size and increases the inference speed. We further apply this dynamic strategy on different distillation temperatures with diverse models to prove the effectiveness and generalization of our method.

{\small
\bibliographystyle{ieee}
\bibliography{egbib}
}

\end{document}